%% file: 0.Main.tex
\documentclass[10pt,twocolumn,letterpaper]{article}

\usepackage{cvpr}      

\usepackage{graphicx}
\usepackage{amsmath}
\usepackage{amssymb}
\usepackage{booktabs}
\usepackage{xcolor}
\usepackage{tikz}
\usepackage[normalem]{ulem}
\usetikzlibrary{positioning}
\usetikzlibrary{calc}
\usepackage[accsupp]{axessibility}  
\usepackage{bbm}
\usepackage{multirow}

\newlength\savewidth
\newcommand{\tablestyle}[2]{\setlength{\tabcolsep}{#1}\renewcommand{\arraystretch}{#2}\centering\footnotesize}

\usepackage[pagebackref,breaklinks,colorlinks]{hyperref}
\usepackage{paralist}

\usepackage[capitalize]{cleveref}
\crefname{section}{Sec.}{Secs.}
\Crefname{section}{Section}{Sections}
\Crefname{table}{Table}{Tables}
\crefname{table}{Tab.}{Tabs.}

\def\cvprPaperID{3469} 
\def\confName{CVPR}
\def\confYear{2023}

\newif\ifdrafting
\draftingtrue 
\newcommand{\customfootnotetext}[2]{{
  \renewcommand{\thefootnote}{#1}
  \footnotetext[0]{#2}}}
  
\ifdrafting
  \newcommand*{\XT}[1]{\textcolor{orange}{[XT: #1]}}
  \newcommand*{\SL}[1]{\textcolor{blue}{[SL: #1]}}
  \newcommand*{\OG}[1]{\textcolor{olive}{[OG: #1]}}
  \newcommand*{\JK}[1]{\textcolor{teal}{[JK: #1]}}
  \newcommand*{\SD}[1]{\textcolor{red}{[SD: #1]}}
\else
  \newcommand*{\XT}[1]{}
  \newcommand*{\SL}[1]{}
  \newcommand*{\OG} [1] {}
  \newcommand*{\JK} [1] {}
  \newcommand*{\SD} [1] {}
  \newcommand{\TODO} [1] {}
\fi

\begin{document}

\title{Zero-shot Pose Transfer for Unrigged Stylized 3D Characters}

\author{Jiashun Wang\textsuperscript{1*} \quad Xueting Li\textsuperscript{2} \quad Sifei Liu\textsuperscript{2} \quad Shalini De Mello\textsuperscript{2} \quad \\
Orazio Gallo\textsuperscript{2} \quad Xiaolong Wang\textsuperscript{3} \quad Jan Kautz\textsuperscript{2}\\
\textsuperscript{1}Carnegie Mellon University \quad \textsuperscript{2}NVIDIA \quad \textsuperscript{3}UC San Diego
}

\twocolumn[{%
\renewcommand\twocolumn[1][]{#1}%
\maketitle
 \begin{center}
     \vspace{-7mm}
     \centering
      \begin{tikzpicture}[inner sep = 0]
          \node(image){\includegraphics[width=\textwidth, trim ={35 35 20 40}, clip]{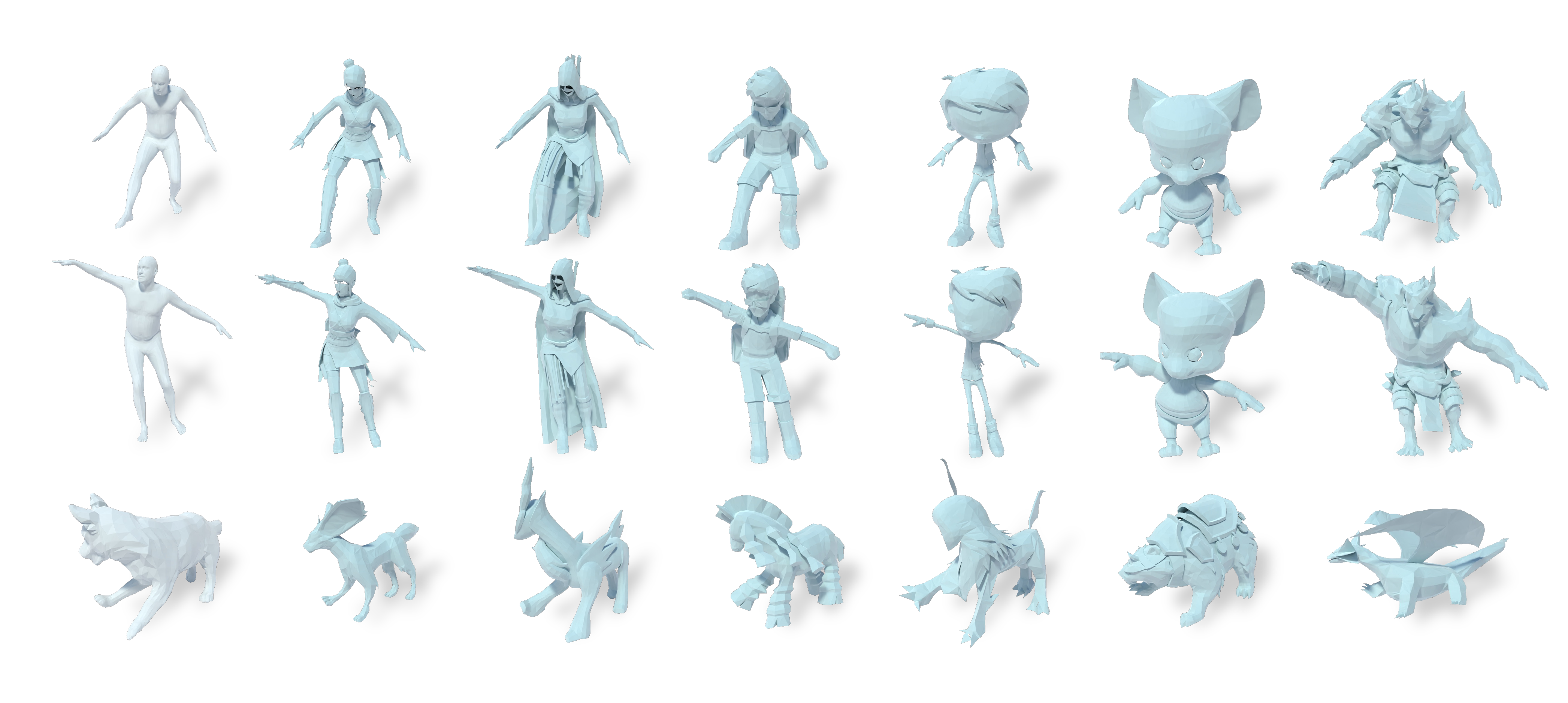}};
          \node(pt1)[below right = 10pt and 0pt of image.south west]{};
          \node(pt2) at ($(pt1)+(65pt,0)$) {};
          \node(pt3) at ($(pt2)+(10pt,0)$) {};
          \node(pt4)[below = 10pt of image.south east]{};
          \draw[lightgray] (pt1)--(pt2);
          \draw[lightgray] (pt3)--(pt4);
          \node(label1)[fill=white,inner sep = 3pt] at($(pt1)!0.5!(pt2)$){Source\strut};
          \node(label2)[fill=white,inner sep = 3pt] at($(pt3)!0.5!(pt4)$){Zero-shot pose transfer onto stylized characters\strut};
     \end{tikzpicture}
     \vspace{-0.2in}
     \captionof{figure}{Our algorithm transfers the pose of a reference avatar (source) to stylized characters. Unlike existing methods, at training time our approach needs only the mesh of the source avatar in rest and desired pose, and the mesh of the stylized characater only in rest pose.}
     \vspace{0.04in}
     \label{fig:teaser}
 \end{center}%
}]

\customfootnotetext{*}{Work done during Jiashun Wang's internship at NVIDIA.}

\input{1.Abstract}
\input{2.Introduction}

\input{3.RelatedWork}
\input{4.Method}
\input{5.Experiments}

\input{6.Conclusion}
\input{8.Appendix}

\clearpage

{\small
\bibliographystyle{ieee_fullname}
\bibliography{egbib}
}

\end{document}

%% file: 1.Abstract.tex

\begin{abstract}
    Transferring the pose of a reference avatar to stylized 3D characters of various shapes is a fundamental task in computer graphics.
    Existing methods either require the stylized characters to be rigged, or they use the stylized character in the desired pose as ground truth at training.
    We present a zero-shot approach that requires only the widely available deformed non-stylized avatars in training, and deforms stylized characters of significantly different shapes at inference.
    Classical methods achieve strong generalization by deforming the mesh at the triangle level, but this requires labelled correspondences.
    We leverage the power of local deformation, but without requiring explicit correspondence labels.
    We introduce a semi-supervised shape-understanding module to bypass the need for explicit correspondences at test time, and an implicit pose deformation module that deforms individual surface points to match the target pose.
    Furthermore, to encourage realistic and accurate deformation of stylized characters, we introduce an efficient volume-based test-time training procedure.
    Because it does not need rigging, nor the deformed stylized character at training time, our model generalizes to categories with scarce annotation, such as stylized quadrupeds.
    Extensive experiments demonstrate the effectiveness of the proposed method compared to the state-of-the-art approaches trained with comparable or more supervision. Our project page is available at \url{https://jiashunwang.github.io/ZPT/}

\end{abstract}







%% file: 2.Introduction.tex
\section{Introduction}
Stylized 3D characters , such as those in Fig.~\ref{fig:teaser}, are commonly used in animation, movies, and video games.
Deforming these characters to mimic natural human or animal poses has been a long-standing task in computer graphics. 
%
Different from the 3D models of natural humans and animals, stylized 3D characters are created by professional artists through imagination and exaggeration.
As a result, each stylized character has a distinct skeleton, shape, mesh topology, and usually include various accessories, such as a cloak or wings (see Fig.~\ref{fig:teaser}).
%
These variations hinder the process of matching the pose of a stylized 3D character to that of a reference avatar, generally making manual rigging a requirement.
Unfortunately, rigging is a tedious process that requires manual effort to create the skeleton and skinning weights for each character.
%
Even when provided with manually annotated rigs, transferring poses from a source avatar onto stylized characters is not trivial when the source and target skeletons differ.
Automating this procedure is still an open research problem and is the focus of many recent works~\cite{poirier2009rig,gleicher1998retargetting,al2018robust,aberman2020skeleton}.
%
Meanwhile, non-stylized 3D humans and animals have been well-studied by numerous prior works~\cite{loper2015smpl,zuffi20173d,li2021learning,ruegg2022barc,wang2020neural}. A few methods generously provide readily available annotated datasets~\cite{zuffi20173d,bhatnagar2019multi,AMASS:2019,bogo2014faust}, or carefully designed parametric models~\cite{loper2015smpl,zuffi20173d,SMPL-X:2019}.
%
By taking advantage of these datasets~\cite{AMASS:2019,bogo2014faust}, several learning-based methods~\cite{wang2020neural,chen2021intrinsic,baran2007automatic,zhou2020unsupervised,li2021learning} disentangle and transfer poses between human meshes using neural networks.
%
However, these methods (referred to as ``part-level'' in the following) carry out pose transfer by either globally deforming the whole body mesh~\cite{chen2021intrinsic,gao2018automatic,zhou2020unsupervised,palafox2021npms} or by transforming body parts~\cite{li2021learning,palafox2022spams}, both of which lead to overfitting on the training human meshes and fail to generalize to stylized characters with significantly different body part shapes.
Interestingly, classical mesh deformation methods~\cite{sumner2004deformation,sumner2007embedded} (referred to as ``local'' in the following) can transfer poses between a pair of meshes with significant shape differences by computing and transferring per-triangle transformations through correspondence.
Though these methods require manual correspondence annotation between the source and target meshes, they provide a key insight that by transforming individual triangles instead of body parts, the mesh deformation methods are more agnostic to a part's shape and can generalize to meshes with different shapes.
%

%
%
We marry the benefits of learning-based methods~\cite{wang2020neural,chen2021intrinsic,baran2007automatic,zhou2020unsupervised,li2021learning} with the classic local deformation approach~\cite{sumner2004deformation} and present a model for unrigged, stylized character deformation guided by a non-stylized biped or quadruped avatar.
Notably, our model only requires easily accessible posed human or animal meshes for training and can be directly applied to deform 3D stylized characters with a significantly different shape at inference.
%
%
%
To this end, we implicitly operationalize the key insight from the local deformation method~\cite{sumner2004deformation} by modeling the shape and pose of a 3D character with a correspondence-aware shape understanding module and an implicit pose deformation module.
The shape understanding module learns to predict the part segmentation label (\ie, the coarse-level correspondence) for each surface point, besides representing the shape of a 3D character as a latent shape code.
The pose deformation module is conditioned on the shape code and deforms individual surface point guided by a target pose code sampled from a prior pose latent space~\cite{pavlakos2019expressive}.
Furthermore, to encourage realistic deformation and generalize to rare poses, we propose a novel volume-based test-time training procedure that can be efficiently applied to unseen stylized characters.
During inference, by mapping biped or quadruped poses from videos, in addition to meshes to the prior pose latent space using existing works~\cite{SMPL-X:2019,kocabas2020vibe,rempe2021humor}, we can transfer poses from different modalities onto unrigged 3D stylized characters.
%
%
Our main contributions are:
\begin{compactitem}
\item We propose a solution to a practical and challenging task -- learning a model for stylized 3D character deformation with only posed human or animal meshes. 
\item  We develop a correspondence-aware shape understanding module, an implicit pose deformation module, and a volume-based test-time training procedure to generalize the proposed model to unseen stylized characters and arbitrary poses in a zero-shot manner. 
\item  We carry out extensive experiments on both humans and quadrupeds to show that our method produces more visually pleasing and accurate deformations compared to baselines trained with comparable or more supervision.
\end{compactitem}

%% file: 3.RelatedWork.tex
\section{Related Work}
\label{sec:related}


\textbf{Deformation Transfer.}
Deformation transfer is a long-standing problem in the computer graphics community~\cite{sumner2004deformation, ben2009spatial, baran2009semantic, avril2016animation, yang2018biharmonic, aigerman2022neural}. Sumner~\etal~\cite{sumner2004deformation} apply an affine transformation to each triangle of the mesh to solve an optimization problem that matches the deformation of the source mesh while maintaining the shape of the target mesh. Ben-Chen~\etal~\cite{ben2009spatial} enclose the source and target shapes with two cages and transfer the Jacobians of the source deformation to the target shape. However, these methods need tedious human efforts to annotate the correspondence between the source and target shapes. More recently, several deep learning methods are developed to solve the deformation transfer task. However, they either require manually providing the correspondence~\cite{yifan2020neural} or cannot generalize~\cite{gao2018automatic, zhou2020unsupervised, chen2021intrinsic} to stylized characters with different shapes. Gao~\etal~\cite{gao2018automatic} propose a VAE-GAN based method to leverage the cycle consistency between the source and target shapes. Nonetheless, it can only work on shapes used in training. Wang~\etal~\cite{wang2020neural} introduce conditional normalization used in style transfer for 3D deformation transfer. But the method is limited to clothed-humans and cannot handle the large shape variations of stylized characters. 

We argue that these learning-based methods cannot generalize to stylized characters because they rely on encoding their global information (\eg, body or parts), which is different from traditional works that focus on local deformation, \eg, the affine transformation applied to each triangle in ~\cite{sumner2004deformation}. Using a neural network to encode the global information easily leads to overfitting. For example, models trained on human meshes cannot generalize to a stylized humanoid character. At the same time, early works only focus on local information and cannot model global information such as correspondence between the source and target shapes, which is why they all need human effort to annotate the correspondence. Our method tries to learn the correspondence and deform locally at the same time.

\textbf{Skeleton-based Pose Transfer.}
Besides mesh deformation transfer, an alternative way to transfer pose is to utilize skeletons.
Motion retargeting is also a common name used for transferring poses from one motion sequence to another.
Gleicher \etal~\cite{gleicher1998retargetting} propose a space-time constrained solver aiming to satisfy the kinematics-level constraints and to preserve the characters' original identity. 
Following works~\cite{lee1999hierarchical, choi2000online, aristidou2011fabrik} try to solve inverse-kinematics or inverse rate control to achieve pose transfer. 
There are also dynamics-based methods~\cite{tak2005physically,al2018robust} that consider physics during the retargeting process. 
Recently, learning-based methods~\cite{villegas2018neural, villegas2021contact, lim2019pmnet, delhaisse2017transfer, jang2018variational} train deep neural networks to predict the transformation of the skeleton. Aberman~\etal~\cite{aberman2020skeleton} propose a pooling-based method to transfer poses between meshes with different skeletons. 

All these works highly rely on the skeleton for pose transfer.
Other works try to estimate the rigging of the template shape~\cite{baran2007automatic, poirier2009rig, xu2019predicting, liu2019neuroskinning, xu2020rignet} when a skeleton is not available. 
But if the prediction of the skinning weights fails, the retargeting fails as well. Liao \etal~\cite{liao2022skeleton} propose a model that learns to predict the skinning weights and pose transfer jointly using ground truth skinning weights and paired motion data as supervision, which limits the generalization of this method to categories where annotations are more scarce compared to humans (\eg, quadrupeds). Instead, our method uses posed human or animal
meshes for training and deforms stylized characters of different shapes at inference.

\textbf{Implicit 3D shape representation.}
Implicit 3D shape representations have shown great success in reconstructing static shapes~\cite{park2019deepsdf, chen2019learning, genova2019learning, mescheder2019occupancy, michalkiewicz2019deep, chibane2020implicit, jiang2020local, erler2020points2surf, chabra2020deep} and deformable ones~\cite{niemeyer2019occupancy, bhatnagar2020combining, palafox2021npms, mihajlovic2021leap, jiang2021learning, palafox2022spams, lei2022cadex,noguchi2022watch, sundararaman2022implicit}. 
 DeepSDF~\cite{park2019deepsdf} proposes to use an MLP to predict the signed distance field (SDF) value of a query point in 3D space, where a shape code is jointly optimized in an auto-decoding manner. 
 Occupancy flow~\cite{niemeyer2019occupancy} generalizes the Occupancy Networks~\cite{mescheder2019occupancy} to learn a temporally and spatially continuous vector field with a NeuralODE~\cite{chen2018neural}.
 %
 Inspired by parameteric models, NPMs~\cite{palafox2021npms} disentangles and represents the shape and pose of dynamic humans by learning an implicit shape and pose function, respectively.
 Different from these implicit shape representation works that focus on reconstructing static or deformable meshes, we further exploit the inherent continuity and locality of implicit functions to deform stylized characters to match a target pose in a zero-shot manner.
 

%% file: 4.Method.tex
\section{Method}
\label{sec:method}

\begin{figure*}[htp]
    \centering
    \vspace{-0.1in}
    \includegraphics[width=0.95\textwidth]{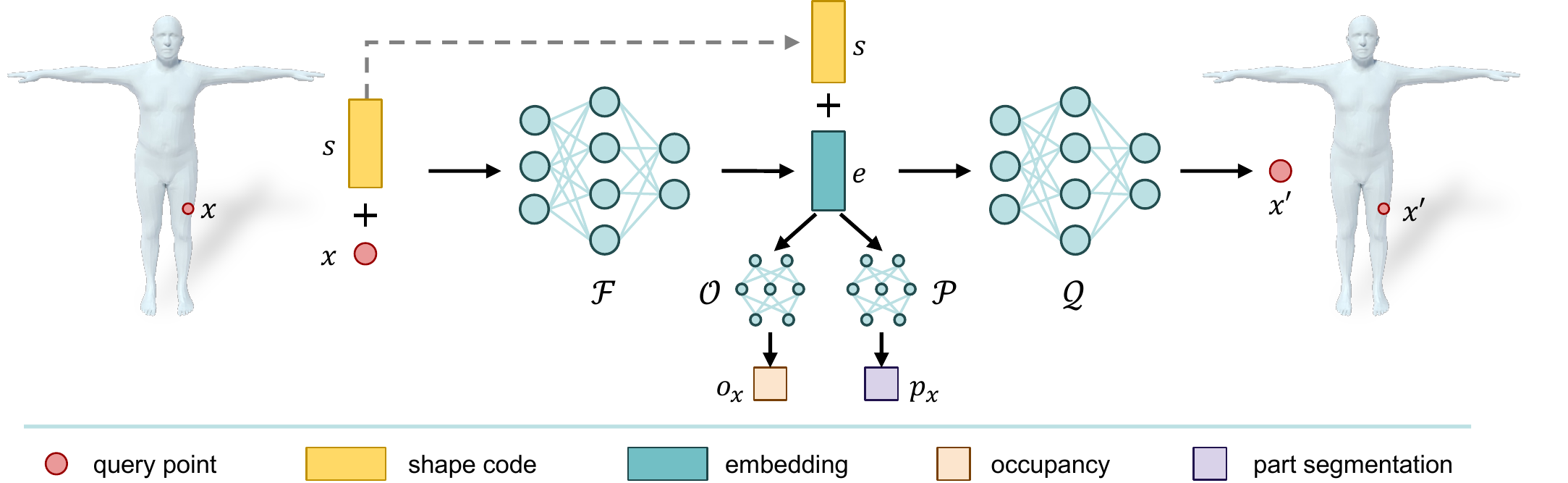}
    \caption{\textbf{The shape understanding module (Sec.~\ref{sec:shape_module}).} Given a query point and a learnable shape code, we take MLPs to predict the occupancy, part segmentation label and further use an inverse MLP to regress the query point.}
    \label{fig::shape_understanding}
\end{figure*}

\begin{figure}[htp]
    \centering
    \vspace{-0.2in}
    \includegraphics[width=0.5\textwidth]{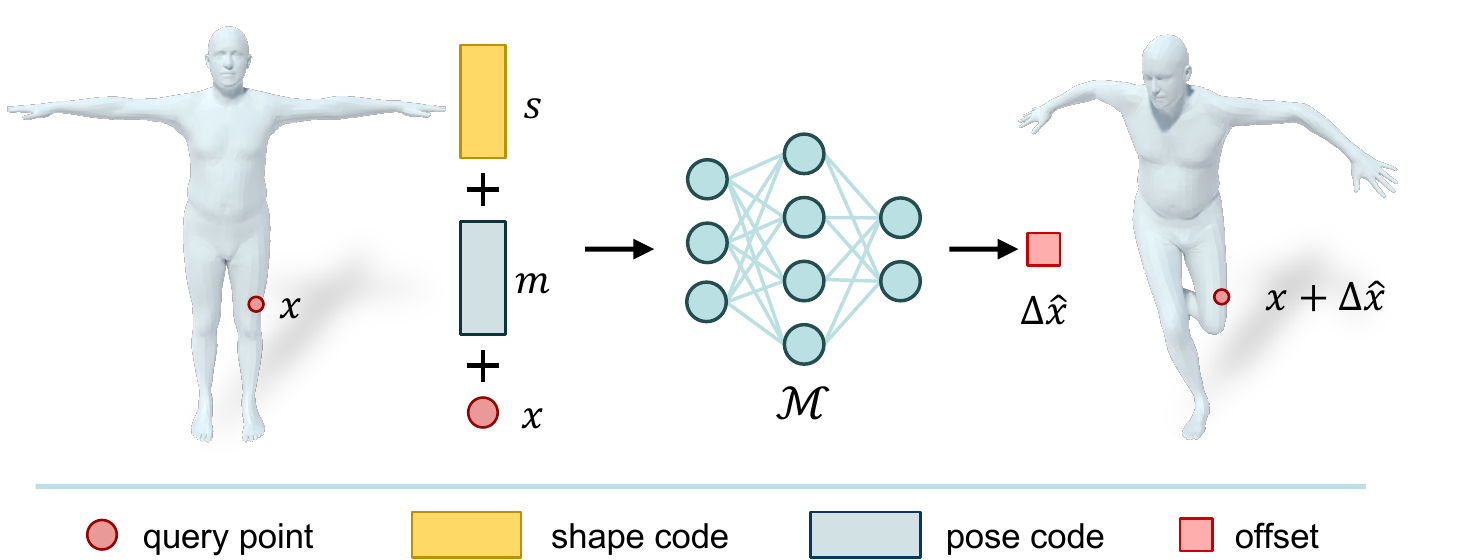}
    \caption{\textbf{The pose deformation module (Sec.~\ref{sec:pose_module}).} Given a query point on the surface, the learned shape code and a target pose code, we use an MLP to predict the offset of the query point.}
    \vspace{-0.1in}
    \label{fig::pose_deformation}
\end{figure}
We aim to transfer the pose of a biped or quadruped avatar to an unrigged, stylized 3D character. 
We tackle this problem by modeling the shape and pose of a 3D character using a correspondence-aware shape understanding module and an implicit pose deformation module, inspired by classical mesh deformation methods~\cite{sumner2004deformation,sumner2007embedded}.
%
%
%
%
The shape understanding module (Sec.~\ref{sec:shape_module}, Fig.~\ref{fig::shape_understanding}) predicts a latent shape code and part segmentation label of a 3D character in rest pose, while the pose deformation module (Sec.~\ref{sec:pose_module}, Fig.~\ref{fig::pose_deformation}) deforms the character in the rest pose given the predicted shape code and a target pose code. 
Moreover, to produce natural deformations and generalize to rare poses unseen at training, we introduce an efficient volume-based test-time training procedure (Sec~\ref{sec:ttt_module}) for unseen stylized characters. 
All three modules, trained only with posed, unclothed human meshes, and unrigged, stylized characters in a rest pose, are directly applied to unseen stylized characters at inference.
We explain our method for humans, and describe how we extend it to quadrupeds in Sec.~\ref{sec:quadruped}.
%


\subsection{Correspondence-Aware Shape Understanding}
\label{sec:shape_module}
Given a 3D character in rest pose, we propose a shape understanding module to represent its shape information as a latent code, and to predict a body part segmentation label for each surface point.
%
%
%
%
%
%

To learn a representative shape code, we employ an implicit auto-decoder~\cite{park2019deepsdf,palafox2021npms} that reconstructs the 3D character taking the shape code as input. 
During training, we jointly optimize the shape code of each training sample and the decoder. 
Given an unseen character (\ie, a stylized 3D character) during inference, we obtain its shape code by freezing the decoder and optimizing the shape code to reconstruct the given character. 
Specifically, as shown in Fig.~\ref{fig::shape_understanding}, given the concatenation of a query point $x \in \mathbb{R}^3$ and the shape code $s \in \mathbb{R}^{d}$, we first obtain an embedding $e \in \mathbb{R}^d$ via an MLP denoted as $\mathcal{F}$. Conditioned on the embedding $e$, the occupancy $\hat{o}_x \in \mathbb{R}$ of $x$ is then predicted by another MLP denoted as $\mathcal{O}$. 
%
The occupancy indicates if the query point $x$ is inside or outside the body surface and can be supervised by the ground truth occupancy as:
\begin{equation}
\vspace{-1mm}
\mathcal{L}_{\mathcal{O}} = -\sum_x(o_x\cdot log(\hat{o}_x) + (1 - o_x)\cdot log(1 - \hat{o}_x)),
\label{eq:shape_rec}
\vspace{-1mm}
\end{equation}
where $o_x$ is the ground truth occupancy at point $x$.

Since our shape code eventually serves as a condition for the pose deformation module, we argue that it should also capture the part correspondence knowledge across different instances, in addition to the shape information (\eg, height, weight, and shape of each body part).
%
This insight has been utilized by early local mesh deformation method~\cite{sumner2004deformation}, which explicitly utilizes correspondence to transfer local transformations between the source and target meshes.
%
Our pose deformation process could also benefit from learning part correspondence.
Take the various headgear, hats, and horns on the stylized characters's heads in Fig.~\ref{fig:teaser} as an example. If these components can be ``understood'' as extensions of the character's heads by their shape codes, they will move smoothly with the character's heads during pose deformation. 
Thus, besides mesh reconstruction, we effectively task our shape understanding module with an additional objective: predicting part-level correspondence instantiated as the part segmentation label.
Specifically, we propose to utilize an MLP $\mathcal{P}$ to additionally predict a part label $p_x=(p^1_x,..., p^K_x)^T \in \mathbb{R}^K $ for each surface point $x$.
Thanks to the densely annotated human mesh dataset, we can also supervise part segmentation learning with ground truth labels via:
\begin{equation}
\mathcal{L}_{\mathcal{P}}= \sum_x (-\sum^{K}_{k=1}\mathbbm{1}^{k}_x log (p^{k}_x)),
\end{equation}
where $K$ is the total number of body parts, and $\mathbbm{1}^{k}_x = 1$ if $x$ belongs to the $k^{th}$ part and $\mathbbm{1}^{k}_x = 0$ otherwise.
To prepare the shape understanding module for stylized characters during inference, besides unclothed human meshes, we also include \textit{unrigged} 3D stylized characters in rest pose during training. These characters in rest pose are easily accessible and do not require any annotation.
For shape reconstruction, Eq.~\ref{eq:shape_rec} can be similarly applied to the stylized characters.
However, as there is no part segmentation annotation for stylized characters, we propose a self-supervised inverse constraint inspired by correspondence learning methods~\cite{liu2020learning,cheng2021learning} to facilitate part segmentation prediction on these characters. 
%
%
Specifically, we reconstruct the query point's coordinates from the concatenation of the shape code $s$ and the embedding $e$ through an MLP $\mathcal{Q}$ and add an auxiliary objective as:
\begin{equation}
\vspace{-1mm}
\label{eq:inverse}
\mathcal{L}_{\mathcal{Q}}= || \mathcal{Q}(s, e) - x||^{2}.
\end{equation}
Intuitively, for stylized characters without part annotation, the model learned without this objective may converge to a trivial solution where similar embeddings are predicted for points with the same occupancy value, even when they are far away from each other, and belong to different body parts. Tab.~\ref{tab::ablation} further quantitatively verifies the effectiveness of this constraint.
Beyond facilitating shape understanding, the predicted part segmentation label is further utilized in the volume-based test-time training module which will be introduced in Sec.~\ref{sec:ttt_module}.

\subsection{Implicit Pose Deformation Module}
\label{sec:pose_module}
Given the learned shape code and a target pose, the pose deformation module deforms each surface point of the character to match the target pose. 
In the following, we first describe how we represent a human pose and then introduce the implicit function used for pose deformation.
%
%
%
%

Instead of learning a latent pose space from scratch as in~\cite{liao2022skeleton,palafox2021npms}, we propose to represent a human pose by the corresponding pose code in the latent space of VPoser~\cite{SMPL-X:2019}.
Our intuition is that VPoser is trained with an abundance of posed humans from the large-scale AMASS dataset~\cite{AMASS:2019}. 
This facilitates faster training and provides robustness to overfitting.
%
Furthermore, human poses can be successfully estimated from different modalities (\eg, videos or meshes), and mapped to the latent space of VPoser by existing methods~\cite{SMPL-X:2019,kocabas2020vibe,rempe2021humor}.
%
By taking advantage of these works, our model can be applied to transfer poses from various modalities to an unrigged stylized character without any additional effort. A few examples can be found in the supplementary.
%

To deform a character to match the given pose, we learn a neural implicit function $\mathcal{M}$ that takes the sampled pose code $m \in \mathbb{R}^{32}$, the learned shape code, and a query point $x$ around the character's surface as inputs and outputs the offset (denoted as $\Delta \hat{x} \in \mathbb{R}^{3}$) of $x$ in 3D space. 
Given the densely annotated human mesh dataset, we directly use the ground truth offset $\Delta x$ as supervision. The training objective for our pose deformation module is defined as:
\begin{align}
\vspace{-1mm}
\mathcal{L}_{\mathcal{D}}= \sum_x || \Delta \hat{x}-\Delta x||^{2}.
\vspace{-1mm}
\end{align}
%

Essentially, our implicit pose deformation module is similar in spirit to early local mesh deformation methods~\cite{sumner2004deformation} and has two key advantages compared to the part-level pose transfer methods~\cite{liao2022skeleton,wang2020neural,gao2018automatic}.
%
%
First, our implicit pose deformation network is agnostic to mesh topology and resolution. Thus our model can be directly applied to unseen 3D stylized characters with significantly different resolutions and mesh topology compared to the training human meshes during inference. 
%
Second, stylized characters often include distinct body part shapes compared to humans. For example, the characters shown in Fig.~\ref{fig:teaser} include big heads or various accessories. Previous part-level methods~\cite{liao2022skeleton} that learn to predict a bone transformation and skinning weight for each body part usually fail on these unique body parts, since they are different from the corresponding human body parts used for training. In contrast, by learning to deform individual surface point, implicit functions are more agnostic to the overall shape of a body part and thus can generalize better to stylized characters with significantly different body part shapes.
%
Fig.~\ref{fig::compare} and Fig.~\ref{fig::compare_animal} show these advantages.




\subsection{Volume-based Test-time Training}
\label{sec:ttt_module}
The shape understanding and pose deformation modules discussed above are trained with only posed human meshes and unrigged 3D stylized characters in rest pose.
When applied to unseen characters with significantly different shapes, we observe surface distortion introduced by the pose deformation module.
Moreover, it is challenging for the module to fully capture the long tail of the pose distribution.
To resolve these issues, we propose to apply test-time training~\cite{sun2020test} and fine-tune the pose deformation module on unseen stylized characters.

To encourage natural pose deformation, we further propose a volume-preserving constraint during test-time training.
%
Our key insight is that preserving the volume of each part in the rest pose mesh during pose deformation results in less distortion~\cite{wang2020neural,li2021learning}.
%
However, it is non-trivial to compute the precise volume of each body part, which can have complex geometry. Instead, we propose to preserve the Euclidean distance between pairs of vertices sampled from the surface of the mesh, as a proxy for constraining the volume.
%
Specifically, given a mesh in rest pose, we randomly sample two points $x^{c}_i$ and $x^{c}_j$ on the surface within the same part $c$ using the part segmentation prediction from the shape understanding module. 
We calculate the offset of these two points $\Delta \hat{x}^{c}_i$ and  $\Delta \hat{x}^{c}_j$ using our pose deformation module and minimize the change in the distance between them by:
\begin{equation}
\vspace{-1mm}
\label{eq: volume}
\mathcal{L}_{v}= \sum_c \sum_i \sum_j (||x^{c}_i-x^{c}_j||-||(x^{c}_i+\Delta \hat{x}^{c}_i)-(x^{c}_j+\Delta \hat{x}^{c}_j)||  )^2.
\end{equation}
By sampling a large number of point pairs within a part and minimizing Eq.~\ref{eq: volume}, we can approximately maintain the volume of each body part during pose deformation. 
%

Furthermore, in order to generalize the pose deformation module to long-tail poses that are rarely seen during training, we propose to utilize the source character in rest pose and its deformed shape as paired training data during test-time training. 
Specifically, we take the source character in rest pose, its target pose code, and its optimized shape code as inputs and we output the movement $\Delta \hat{x}^{dr}$, where $x^{dr}$ is a query point from the source character. We minimize the L2 distance between the predicted movement $\Delta \hat{x}^{dr}$ and the ground truth movement $\Delta x^{dr}$,
\begin{equation}
\vspace{-1mm}
\mathcal{L}_{dr}= \sum_{x^{dr}} || \Delta \hat{x}^{dr}-\Delta x^{dr}||^{2}.
\vspace{-1mm}
\end{equation}
Besides the volume-preserving constraint and the reconstruction of the source character, we also employ the edge loss $\mathcal{L}_{e}$ used in ~\cite{groueix20183d,wang2020neural,liao2022skeleton}. Overall, the objectives for the test-time training procedure are $\mathcal{L}_{\mathcal{T}}= \lambda_v \mathcal{L}_{v} + \lambda_e\mathcal{L}_{e} + \lambda_{dr} \mathcal{L}_{dr}$,
where $\lambda_v$, $\lambda_e$, and $\lambda_{dr}$ are hyper-parameters balancing the loss weights. 
%

%

%% file: 5.Experiments.tex
\section{Experiments}
\label{sec:exp}

\subsection{Datasets}
To train the shape understanding module, we use 40 human meshes sampled from the SMPL~\cite{loper2015smpl} parametric model. 
We use both the occupancy and part segmentation label of these meshes as supervision (see Sec.~\ref{sec:shape_module}).
To generalize the shape understanding module to stylized characters, we further include 600 stylized characters from RigNet~\cite{xu2020rignet}. 
Note that we \textit{only} use the rest pose mesh (\ie, occupancy label) of the characters in~\cite{xu2020rignet} for training.
%
%
To train our pose deformation module, we construct paired training data by deforming each of the 40 SMPL characters discussed above with 5000 pose codes sampled from the VPoser's~\cite{pavlakos2019expressive} latent space. In total, we collect 200,000 training pairs, with each pair including an unclothed human mesh in rest pose and the same human mesh in target pose.

After training the shape understanding and pose deformation modules, we test them on the Mixamo~\cite{Mixamo} dataset, which includes challenging stylized characters, and the MGN~\cite{bhatnagar2019multi} dataset, which includes clothed humans.
The characters in both datasets have different shapes compared to the unclothed SMPL meshes we used for training, demonstrating the generalization ability of the proposed method.
Following~\cite{liao2022skeleton}, we test on 19 stylized characters, with each deformed by 28 motion sequences from the Mixamo dataset. 
For the MGN dataset, we test on 16 clothed characters, with each deformed by 200 target poses. Both the testing characters and poses are unseen during training. 

For quadrupeds, since there is no dataset including large-scale paired stylized quadrupeds for quantitative evaluation, we split all characters from the SMAL~\cite{zuffi20173d} dataset and use the first 34 shapes (\ie, cats, dogs, and horses) for training.
We further collect 81 stylized quadrupeds in rest pose from the RigNet~\cite{xu2020rignet} to improve generalization of the shape understanding module.
Similarly to the human category, we use occupancy and part segmentation supervision for the SMAL shapes and only the occupancy supervision for RigNet meshes. 
To train the pose deformation module, we deform each of the 34 characters in SMAL by 2000 poses sampled from the latent space of BARC~\cite{ruegg2022barc}, a 3D reconstruction model trained for the dog category.
We quantitatively evaluate our model on the hippo meshes from the SMAL dataset, which have larger shape variance compared to the cats, dogs, and horses used for training. 
We produce the testing data by deforming each hippo mesh with 500 unseen target poses from SMAL~\cite{zuffi20173d}.
We show qualitative pose transfer on stylized quadrupeds in Fig.~\ref{fig:teaser}.

\subsection{Implementation Details}
We use the ADAM~\cite{kingma2014adam} optimizer to train both the shape understanding and pose deformation modules. For the shape understanding module, we use a learning rate of $1e-4$ for both the decoder and shape code optimization, with a batch size of 64. Given a new character at inference time, we fix the decoder and only optimize the shape code for the new character with the same optimizer and learning rate. For the pose deformation module, we use a learning rate of $3e-4$ with a batch size of 128. For test-time training, we use a batch size of 1 and a learning rate of $5e-3$ with the ADAM optimizer. We set $\lambda_v$, $\lambda_e$, and $\lambda_{dr}$ (See Sec.~\ref{sec:ttt_module}) as 0.05, 0.01, and 1 respectively. 


\begin{figure*}
  \centering
  \vspace{-0.2in}
  \input{figures/results/result_figure.tex}
  \vspace{-0.1in}
  \caption{\textbf{Qualitative comparison on Mixamo.} The average PMD of these three results for NBS, SPT, and Ours are 8.16, 6.13, and 5.16 respectively and the average ELS for NBS, SPT, and Ours are 0.65, 0.78, and 0.93 respectively. Our method can successfully transfer the pose to challenging stylized characters (e.g., the mouse with a big head in the second row).}
  \label{fig::compare}
  \vspace{-0.05in}
\end{figure*}
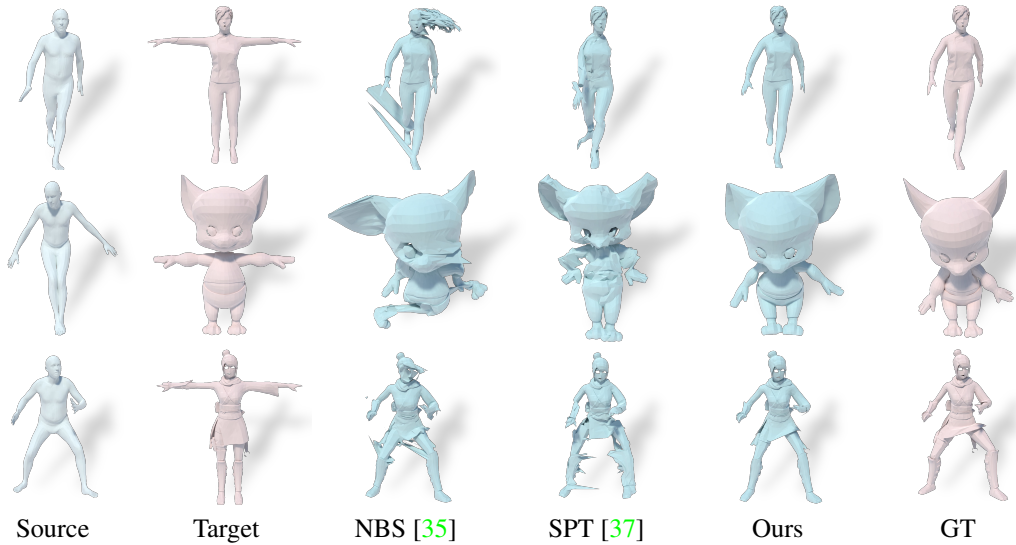


\subsection{Metrics and Baselines for Comparison}
\textbf{Metrics.} We use Point-wise Mesh Euclidean Distance (PMD)~\cite{wang2020neural,liao2022skeleton} to evaluate pose transfer error. The PMD metric reveals pose similarity of the predicted deformation compared to its ground truth. However, as shown in Fig.~\ref{fig::compare}, PMD can not fully show the smoothness and realism of the generated results. Thus, we adopt an edge length score (ELS) metric to evaluate the character's smoothness after the deformation. Specifically, we compare each edge's length in the deformed mesh with the corresponding edge's length in the ground truth mesh. We define the score as
\begin{equation}
\vspace{-1mm}
    \frac{1}{|\mathcal{E}|}\sum_{\{{i,j\} \sim \mathcal{E}}} 1-|\frac{ ||\hat{V_i}-\hat{V_j}||_2}{|| V_i-V_j||_2}-1|,
\vspace{-1mm}
\end{equation}
where $\mathcal{E}$ indicates all edges of the mesh, $|\mathcal{E}|$ is the number of the edges in the mesh. $\hat{V}_i$ and $\hat{V}_j$ are the vertices in the deformed mesh. $V_i$ and $V_j$ are the vertices in the ground truth mesh. For all the evaluation metrics, we scale the template character to be 1 meter tall, following~\cite{liao2022skeleton}.

\textbf{Baselines.} We compare our method with Neural Blend Shapes (NBS)~\cite{li2021learning} and Skeleton-free Pose Transfer (SPT)~\cite{liao2022skeleton}. NBS is a rigging prediction method trained on the SMPL and MGN datasets, which include naked and clothed human meshes with ground truth rigging information. 
For SPT, we show the results of two versions, one is trained only on the AMASS dataset, named SPT, which has a comparable level of supervision to our method. We also test the SPT*(full) version, which is trained on the AMASS, RigNet and Mixamo datasets, using both stylized characters' skinning weights as supervision and paired stylized characters in rest pose and target pose. 
%


\begin{table}
\centering
\tablestyle{3.5pt}{1}
\footnotesize
\begin{tabular}{l l c c c c}
\hline
\hline
Dataset              & Metric  & SPT*(full)~\cite{liao2022skeleton} & NBS~\cite{li2021learning} & SPT~\cite{liao2022skeleton} & Ours\\ \hline
\multirow{2}{*}{MGN~\cite{bhatnagar2019multi}} & PMD $\downarrow$    & 1.62  & 1.33 & 1.82 &  0.99 \\ 
                     & ELS $\uparrow$   & 0.86 & 0.70  & 0.85 &  0.89 \\ \hline
\multirow{2}{*}{Mixamo~\cite{Mixamo}} & PMD $\downarrow$   & 3.05  & 7.04 & 5.29 & 5.06 \\ 
                     & ELS $\uparrow$   & 0.61 & 0.66 & 0.59 & 0.88\\ \hline

\end{tabular}
  \caption{\textbf{Quantitative comparison on MGN and Mixamo.} Our method achieves the lowest PMD with the highest ELS. We provide the performance of the SPT*(full) method, which uses more supervision than the other methods as a reference. Our method is even better or comparable to it. }
  \label{tab::main_comparison}
  \vspace{-0.2in}
\end{table}

\subsection{Human-like Character Pose Transfer}
We report the PMD metric on the MGN and Mixamo datasets in Tab.~\ref{tab::main_comparison}. We also include the performance of SPT*(full) for reference. 
On the MGN dataset which includes clothed humans, our method which is trained with only unclothed humans achieve the best PMD score than all baseline methods, including baselines trained with more supervision (\ie, the NBS~\cite{li2021learning} learned with clothed humans and the SPT*(full)~\cite{liao2022skeleton} learned with skinning weight and paired motion data). 
For the stylized characters, our method outperforms the SPT baseline learned with a comparable amount of supervision and gets competitive results with the NBS~\cite{li2021learning} and SPT*(full) baseline trained with more supervision. 
Furthermore, when testing on the more challenging, less human-like characters (\eg, a mouse with a big head in Fig.~\ref{fig:teaser}), the baselines produce noticeable artifacts and rough surfaces, which can be observed in the qualitative comparisons in Fig.~\ref{fig::compare}. 
We provide the PMD value for each character in the supplementary.

We show the ELS score comparison of different methods on the MGN and Mixamo datasets in Tab.~\ref{tab::main_comparison}. 
%
%
For both clothed humans and stylized characters, our method can generate more realistic results which are consistent with the target mesh and achieves the best ELS score.


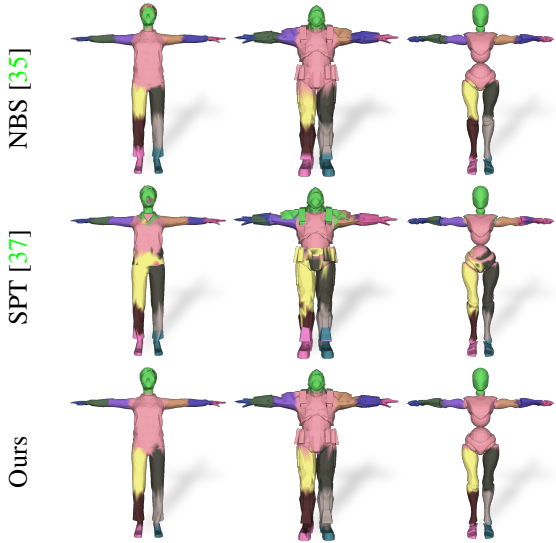
\begin{figure}[t]
  \centering
  \vspace{-0.2in}
  \begin{tikzpicture}
    \node(col_1){\includegraphics[height=3.in, trim={70 20 60 35}, clip]{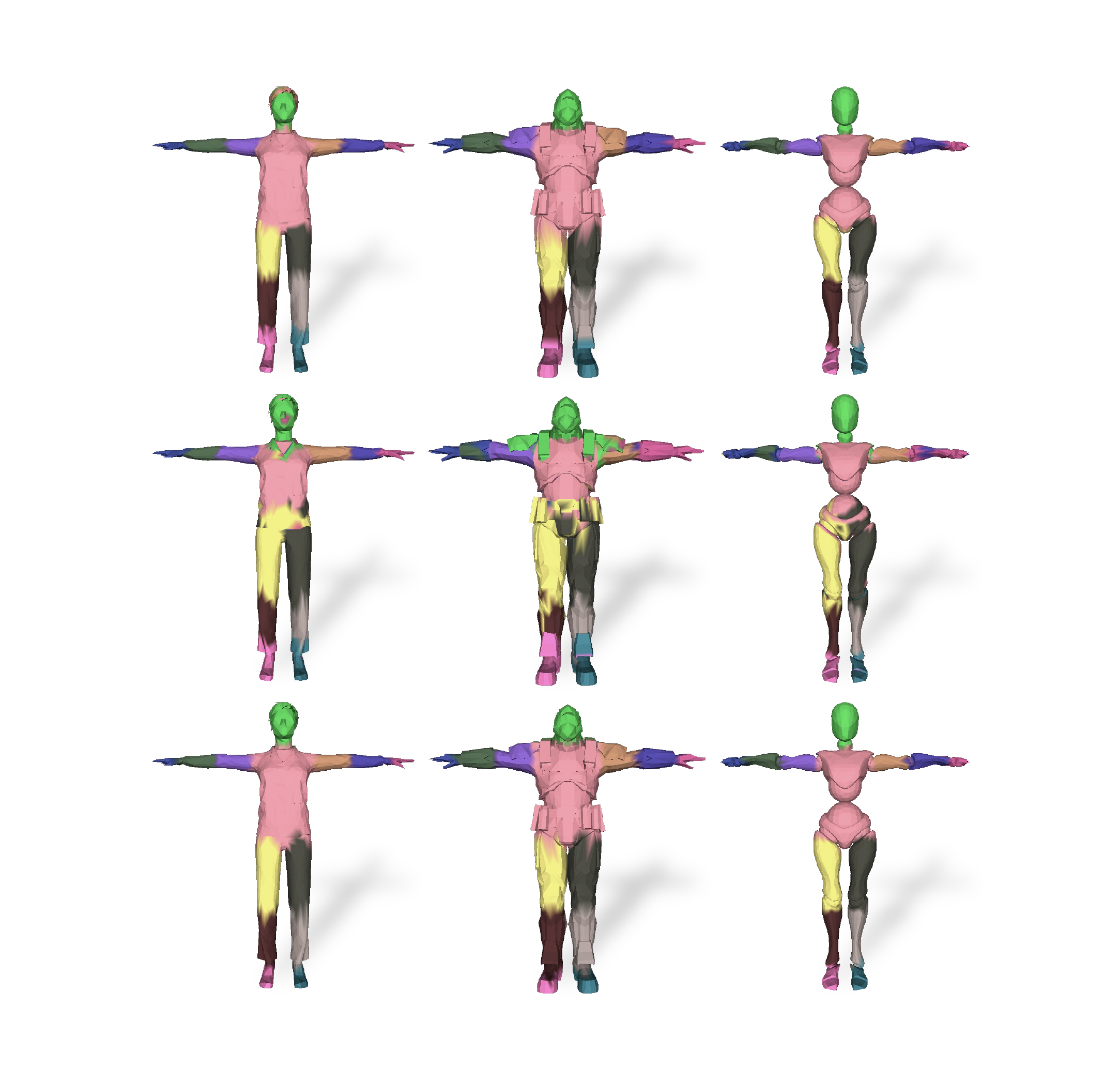}};
    \node(row_2_label)[left = 0pt of col_1, rotate=90, anchor=south]{SPT~\cite{liao2022skeleton}\strut};
    \node(row_1_label)[rotate=90, anchor=west] at ($(row_2_label.east)+(0,23pt)$){NBS~\cite{li2021learning}\strut};
    \node(row_3_label)[rotate=90, anchor=east] at ($(row_2_label.west)-(0,35pt)$){Ours\strut};
  \end{tikzpicture}
  \vspace{-0.25in}
  \caption{\textbf{Part segmentation visualization.} NBS makes wrong predictions for hair while SPT may mix the upper legs.}
  \vspace{-3mm}
  \label{fig::part}
\end{figure}

We visually compare our method and the baseline methods in Fig.~\ref{fig::compare} on the Mixamo dataset. Although NBS is trained with a clothed-human dataset, when testing on the human-like characters, it still fails on parts that are separate from the body such as the hair and the pants. When using only naked human meshes as supervision, SPT cannot generalize to challenging human-like characters, producing rough mesh surface with spikes.

\begin{figure}
  \centering
  \vspace{-5mm}
  \begin{tikzpicture}
      \node(image) at (0,0){\includegraphics[width=0.95\columnwidth,trim={40 20 10 30}, clip]{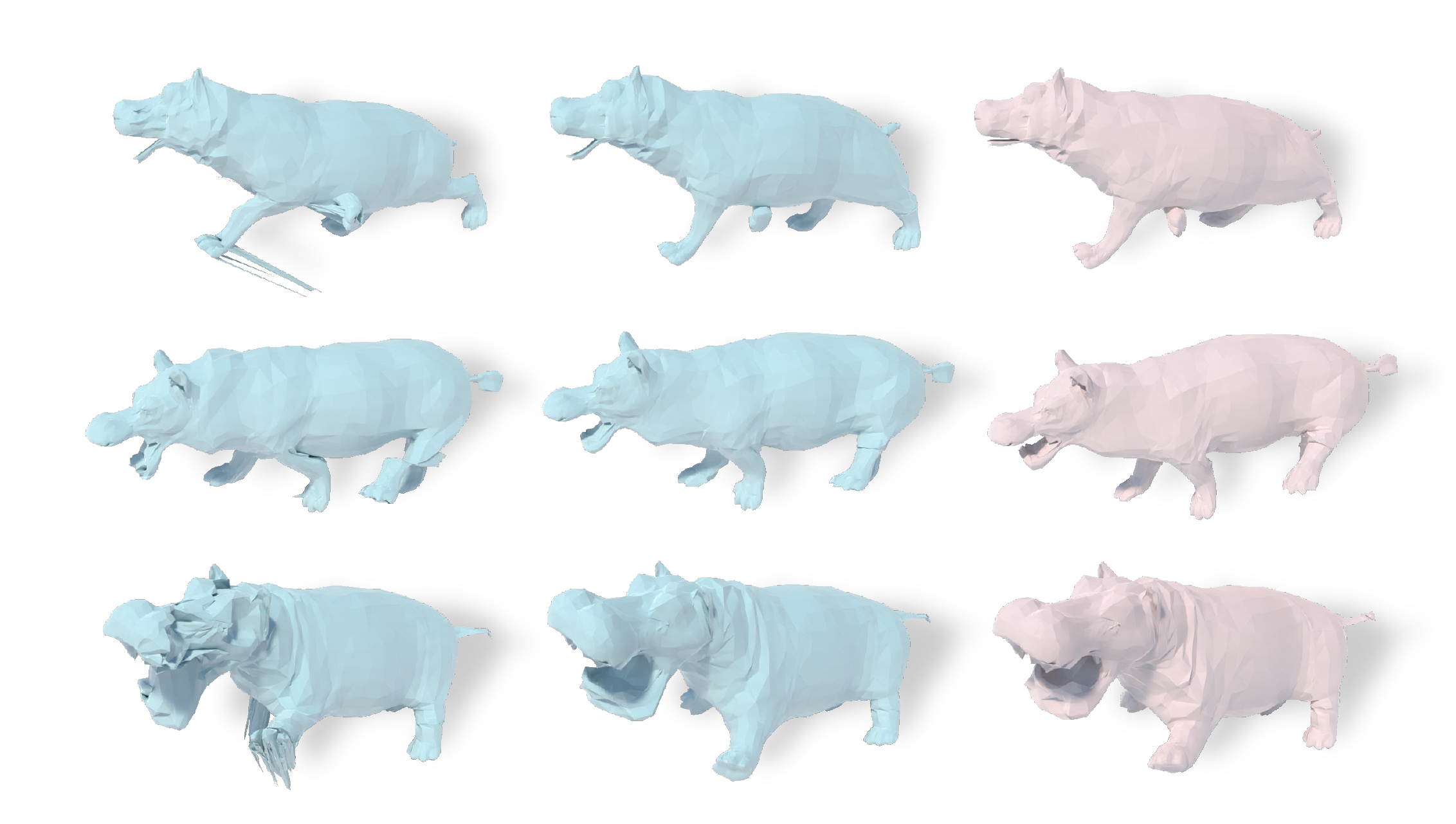}};
      \node(ours_label) [below = 0 of image.south]{Ours\strut};
      \node(spt_label)  [below left = 0 and 55pt of image.south]{SPT~\cite{liao2022skeleton}\strut};
      \node(gt_label) [below right = 0 and 65pt of image.south]{GT\strut};
  \end{tikzpicture}
  \vspace{-3mm}
  \caption{\textbf{Quadrupedal pose transfer visualization.} Our method can achieve smooth and accurate pose transfer while SPT fails on the mouth and leg regions.}
  \label{fig::compare_animal}
\end{figure}
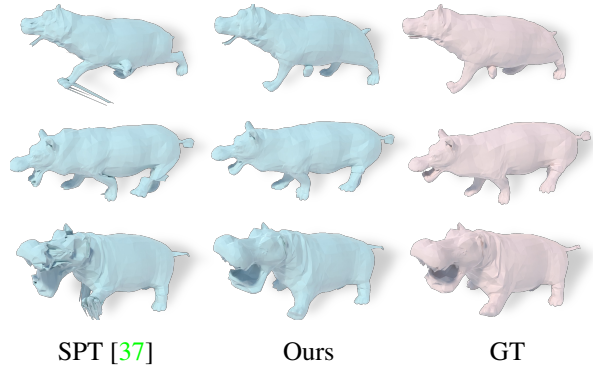

\begin{table}[h]
    \centering
    \footnotesize
    \begin{tabular}{l  c c c} 
        \hline
        \hline
       Metric  & NBS~\cite{li2021learning} & SPT~\cite{liao2022skeleton} & Ours \\ 
       \hline
       Accuracy $\uparrow$ & 67.8\% & 75.6\% & 86.9\% \\
       \hline
    \end{tabular}
    \caption{\textbf{Part prediction accuracy on Mixamo~\cite{Mixamo}}. Our method achieves the best part segmentation accuracy.}
    \label{tab::part}
    \vspace{-2mm}
\end{table}

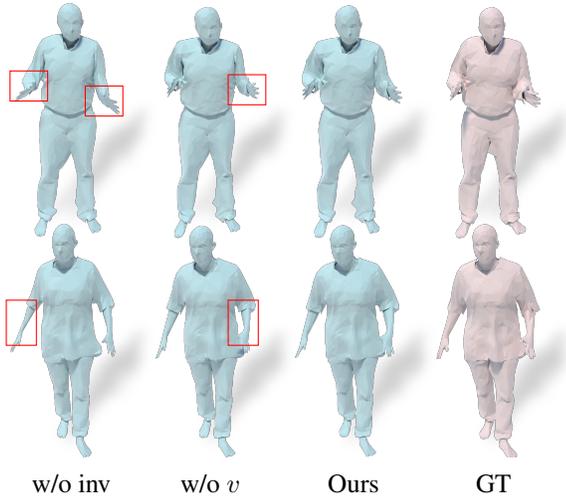
\begin{figure}[!t]
  \centering
  \vspace{-0.2in}
  \begin{tikzpicture}
      \node(image) at (0,0){\includegraphics[width=0.9\columnwidth, trim={85 40 72 25}, clip]{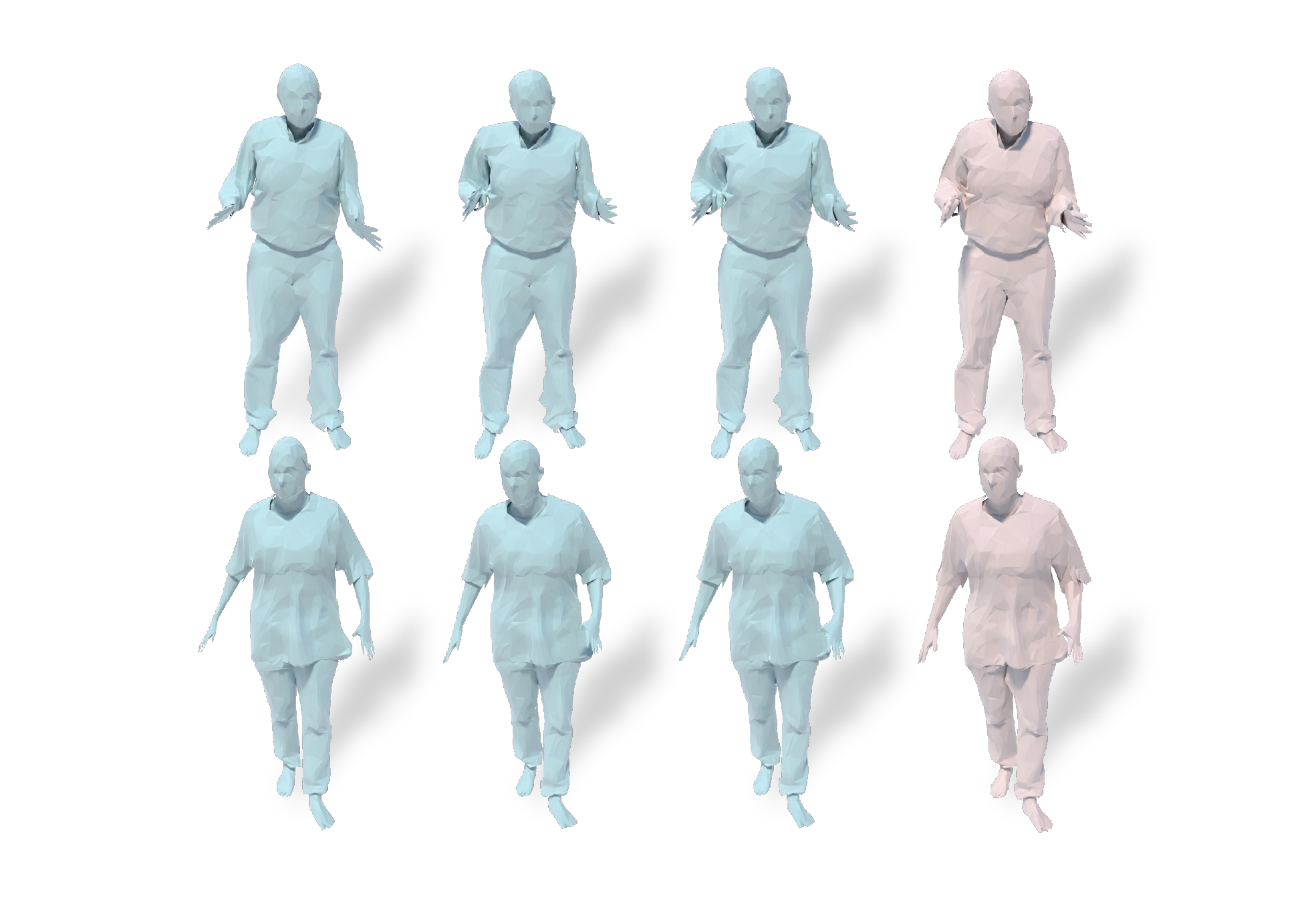}};
      \node(noinv_label) [below left = -2pt and 65pt of image.south]{w/o inv\strut};
      \node(w/o_label)  [below left = -2pt and 16pt of image.south]{ w/o $v$\strut};
      \node(nov_label) [below right = -2pt and 10pt of image.south]{Ours\strut};
      \node(gt_label) [below right = -2pt and 66pt of image.south]{GT\strut};
      \node(err1)[draw, red, minimum width=5mm, minimum height=4mm] at (-3.5 , 1.9){};
      \node(err1)[draw, red, minimum width=5mm, minimum height=4mm] at (-2.5 , 1.7){};

      \node(err1)[draw, red, minimum width=5mm, minimum height=4mm] at (-0.6 , 1.85){};

      \node(err1)[draw, red, minimum width=4mm, minimum height=6mm] at (-3.6 , -1.25){};

      \node(err1)[draw, red, minimum width=4mm, minimum height=6mm] at (-0.65 , -1.25){};
      
  \end{tikzpicture}
  \vspace{-4mm}
  \caption{\textbf{Qualitative comparison for ablation study.} Removing the constraint (eq.~\ref{eq:shape_rec}) in shape understanding leads to wrong pose deformation results. The volume preserving loss (eq.~\ref{eq: volume}) helps to maintain the identity, \eg, the thickness of the arms in first row.}
  \label{fig::ablation}
  \vspace{-1mm}
\end{figure}

\subsection{Part Understanding Comparison}
As discussed in Sec.~\ref{sec:shape_module}, part segmentation plays an important role in both shape understanding and pose deformation.
Though NBS~\cite{li2021learning} and SPT~\cite{liao2022skeleton} do not explicitly predict part segmentation label, they are both skinning weight-based methods and we can derive the part segmentation label from the predicted skinning weights. 
%
%
Specifically, by selecting the maximum weight of each vertex, we can convert the skinning weight prediction to part segmentation labels for the vertices. 
We compare our part prediction results with those derived from SPT and NBS. 
We report the part segmentation accuracy on the Mixamo datasets in Tab.~\ref{tab::part} and visualize the part segmentation results in Fig.~\ref{fig::part}. 
Even trained with only part segmentation supervision of human meshes, our method can successfully segment each part for the stylized characters. 
On the contrary, SPT uses graph convolution network~\cite{kipf2016semi} to predict the skinning weights.
When training only with human meshes, it often fails to distinguish different parts. 
As shown in Fig.~\ref{fig::part}, it mixes up the right and left upper legs, and incorrectly classifies the shoulder as the head. 
Though NBS is trained with clothed humans, it always classifies human hair as the human body for characters from Mixamo. 
This is because that NBS uses the MeshCNN~\cite{hanocka2019meshcnn} as the shape encoder. As a result, it is sensitive to mesh topology and cannot generalize to meshes with disconnected parts (\eg, disconnected hair and head). 
%
Tab.~\ref{tab::part} further quantitatively demonstrates that our method achieves the best part segmentation accuracy, demonstrating its ability to correctly interpret the shape and part information in stylized characters.

\begin{table}
    \centering
    \footnotesize
    \begin{tabular}{l c c |c c c c } 
        \hline
        \hline
        Metric & SPT~\cite{liao2022skeleton} & Ours & Metric & SPT~\cite{liao2022skeleton} & Ours\\ 
       \hline
       PMD $\downarrow$ & 10.28 & 8.28 & ELS $\uparrow$ & 0.28 & 0.86 \\ \hline
    \end{tabular}
    \vspace{-1mm}
    \caption{\textbf{Comparison on Hippos from SMAL~\cite{zuffi20173d}}. Our method achieves better pose transfer accuracy with more smooth results.}
    \label{tab::animal}
    \vspace{-5mm}
\end{table}

\subsection{Quadrupedal Pose Transfer Comparison}
\label{sec:quadruped}

\begin{figure}
  \centering
  \vspace{-0.2in}
  \begin{tikzpicture}
      \node(image) at (0,0){\includegraphics[width=0.95\columnwidth, trim={110 70 80 90}, clip]{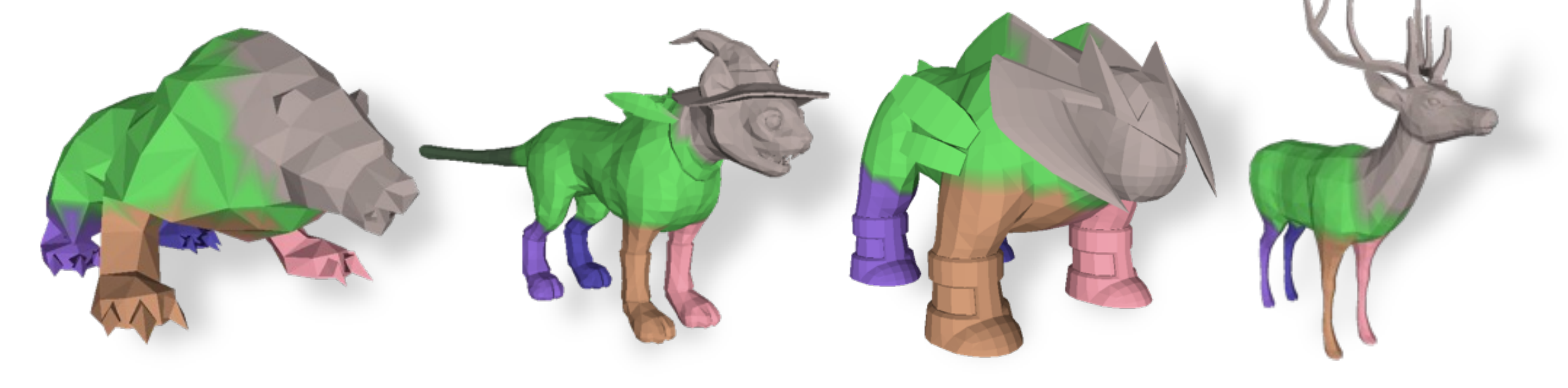}};
  \end{tikzpicture}
  \vspace{-0.15in}
  \caption{\textbf{Part prediction on stylized quadrupeds.} Our method successfully predicts the parts of unseen stylized quadrupeds.}
  \label{fig::animal_part}
  \vspace{-0.05in}
\end{figure}

To further show the generalization ability of our method, we conduct experiments on quadrupeds. 
%
We report the PMD and ELS score of our method and the SPT~\cite{liao2022skeleton} in Tab.~\ref{tab::animal}.
When testing on hippos with large shape gap from the training meshes, SPT has a hard time generalizing both in terms of pose transfer accuracy and natural deformation. 
While our method achieves both better qualitative and quantitative results.  
We visualize the qualitative comparisons in Fig.~\ref{fig::compare_animal}. 
SPT produces obvious artifacts on the hippo's mouth and legs, while our method achieves accurate pose transfer and maintains the shape characteristics of the original character at the same time. 
We provide more results in the supplementary. 
We also show the part segmentation results on stylized characters by our method in Fig.~\ref{fig::animal_part}. 
Even for unique parts such as the hats and antlers, our method correctly assigns them to the head part.

\begin{table}
    \centering
    \footnotesize
    \begin{tabular}{l c c c c c} 
        \hline
        \hline
       Metric  & Ours w/o inv & Ours w/o volume & Ours \\ 
       \hline
        PMD $\downarrow$  & 1.26 & 1.02 & 0.99 \\   \hline
        ELS $\uparrow$ & 0.88  & 0.88 & 0.89 \\ \hline
       
    \end{tabular}
    \vspace{-0.05in}
    \caption{\textbf{Ablation study on inverse MLP and volume preserving loss.} The inverse MLP and volume preserving loss helps to improve pose transfer accuracy and produce smooth deformation.}
    \label{tab::ablation}
    \vspace{-5mm}
\end{table}

\subsection{Ablation Study}
To evaluate the key components of our method, we conduct ablation studies on the MGN dataset by removing the inverse constraint (Eq.~\ref{eq:inverse}) in the shape understanding module and the volume-preserving loss (Eq.~\ref{eq: volume}) used during the test-time training produce, we name them as ``ours w/o inv" and ``ours w/o $v$" respectively. 
We report the PMD and ELS metrics in Tab.~\ref{tab::ablation}.
The model learned without the inverse constraint or volume-preserving loss has worse PMD and ELS score than our full model, indicating the contribution of these two objectives.
%
%
We also provide qualitative results in Fig.~\ref{fig::ablation}. We use red boxes to point out the artifacts. As shown in Fig.~\ref{fig::ablation}, our model trained without the inverse constraint produces less accurate pose transfer results.
Moreover, adding the volume-preserving loss helps to maintain the character's local details such as the thickness of the arms.

%% file: figures/results/result_figure.tex
\begin{tikzpicture}[outer sep = 0, inner sep = 0]
    \newlength{\imgheight}
    \setlength{\imgheight}{0.9in}
    
    \newlength{\colOne}
    \setlength{\colOne}{0pt}
    \newlength{\colTwo}
    \setlength{\colTwo}{10pt}
    \newlength{\colThr}
    \setlength{\colThr}{20pt}
    \newlength{\colFou}
    \setlength{\colFou}{20pt}
    \newlength{\colFiv}
    \setlength{\colFiv}{20pt}
    \newlength{\vgap}
    \setlength{\vgap}{0pt}

    \newlength{\offset}
    \setlength{\offset}{-20pt}

    \node(top_left) at (0,0){};
    \node(r1c1)[anchor=north west] at (top_left){\includegraphics[height=\imgheight, trim = {80 18 85 20}, clip]{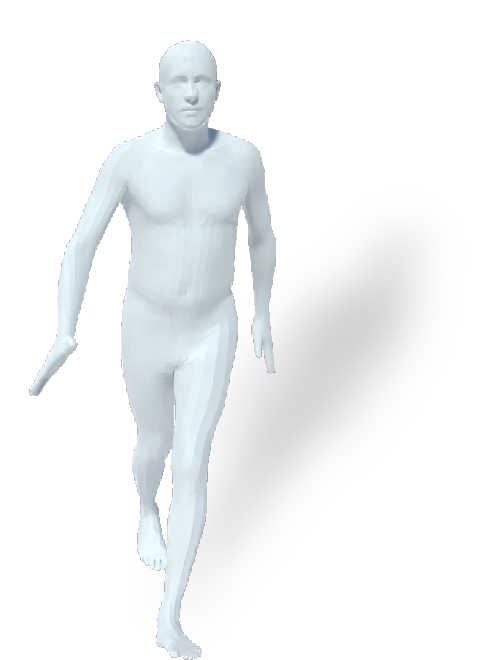}};
    \node(r1c2)[right = \colOne of r1c1]{\includegraphics[height=\imgheight, trim = {30 16 60 20}, clip]{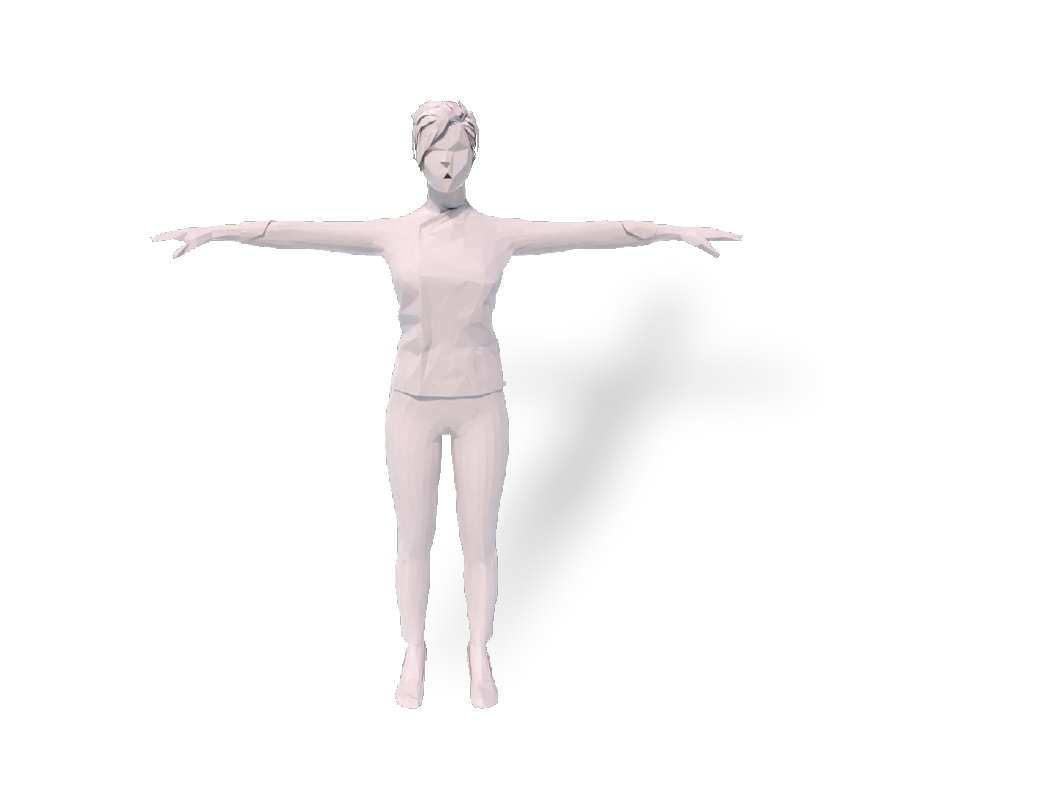}};
    \node(r1c3)[right = \colTwo of r1c2]{\includegraphics[height=\imgheight, trim = {60 16 80 20}, clip]{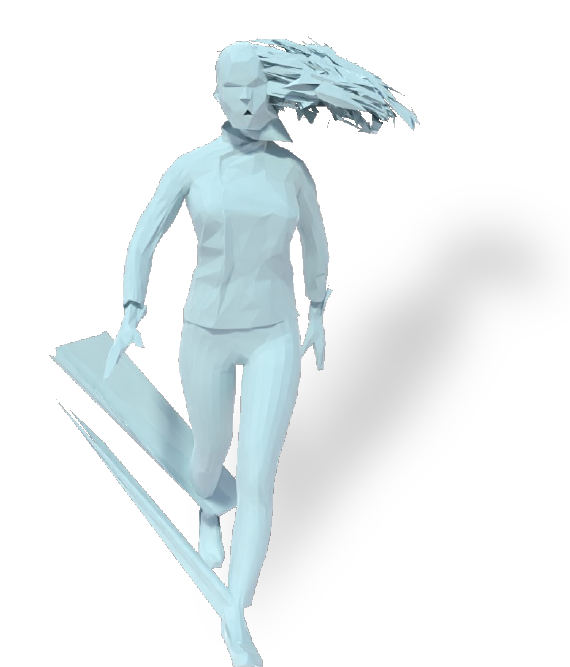}};
    \node(r1c4)[right = \colThr of r1c3]{\includegraphics[height=\imgheight, trim = {80 16 80 20}, clip]{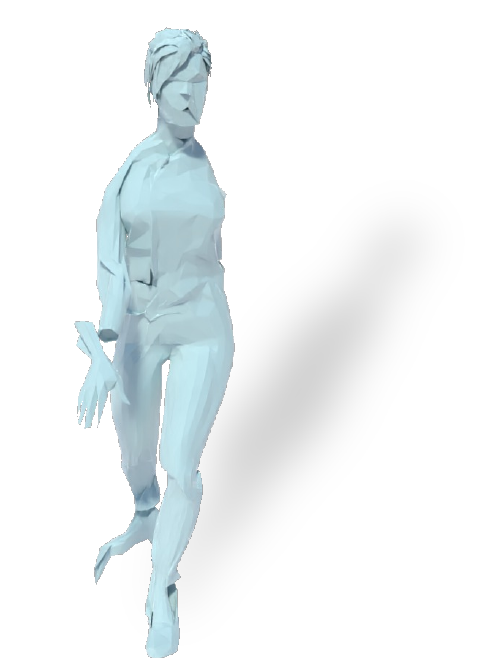}};
    \node(r1c5)[right = \colFou of r1c4]{\includegraphics[height=\imgheight, trim = {80 12 80 22}, clip]{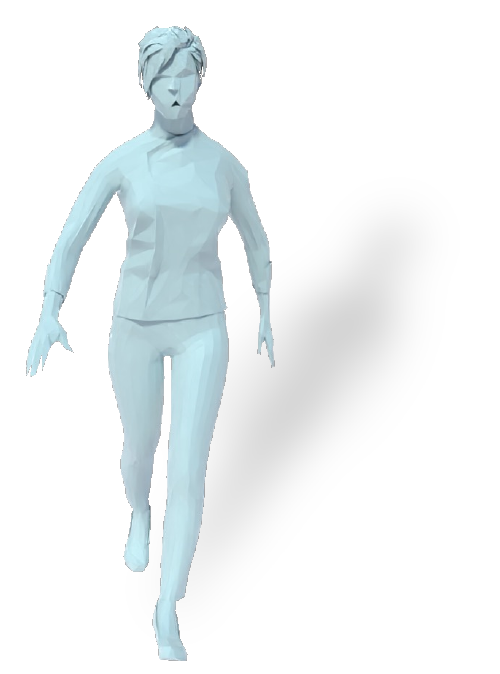}};
    \node(r1c6)[right = \colFiv of r1c5]{\includegraphics[height=\imgheight, trim = {80 12 80 22}, clip]{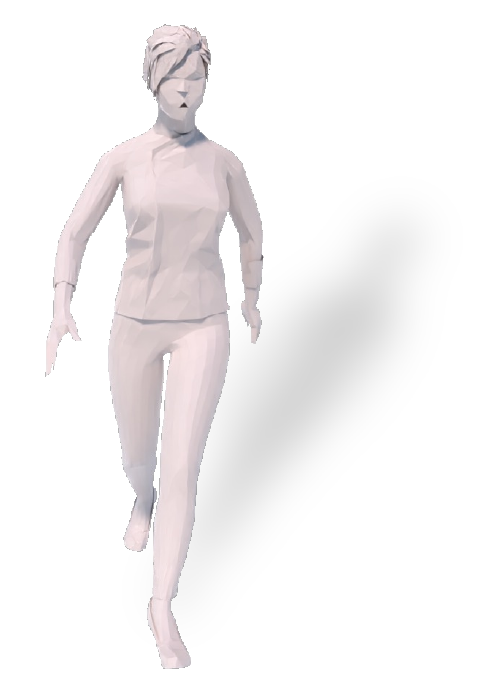}};
    \node(offset_point) at ($(r1c1.north west)+(\offset,0)$){};

    \node(r2c1)[below = \vgap of r1c1]{\includegraphics[height=\imgheight, trim = {60 10 80 25}, clip]{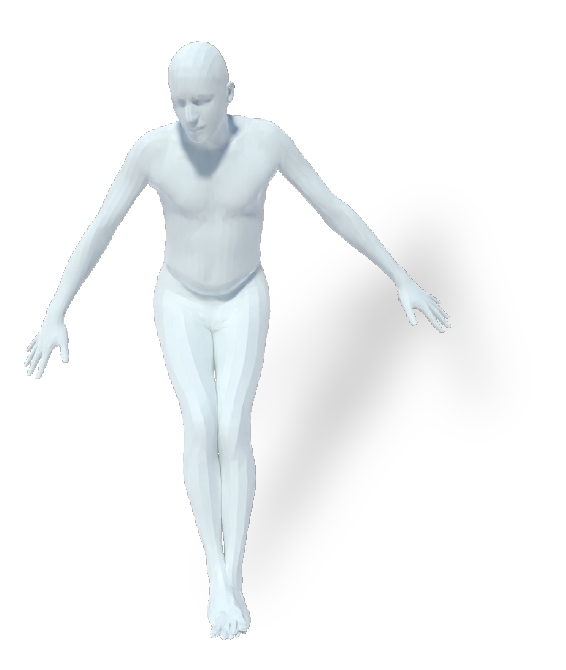}};
    \node(r2c2)[below = \vgap of r1c2]{\includegraphics[height=\imgheight, trim = {40 10 70 25}, clip]{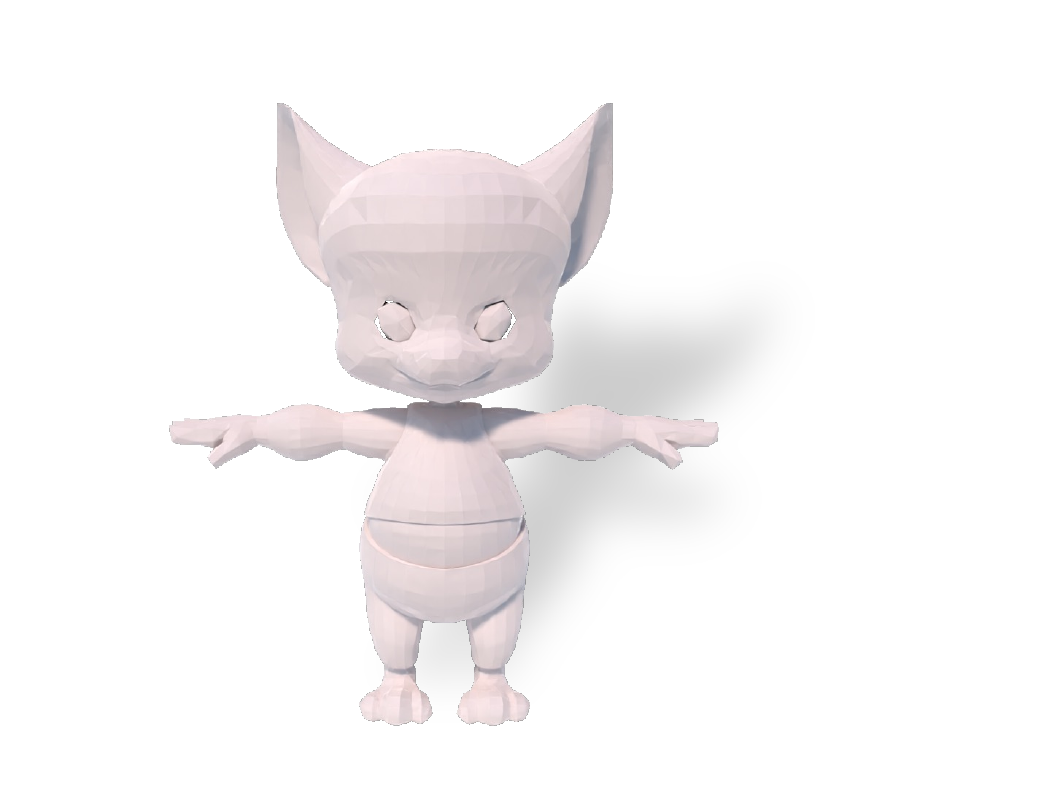}};
    \node(r2c3)[below = \vgap of r1c3]{\includegraphics[height=\imgheight, trim = {24 12 70 32}, clip]{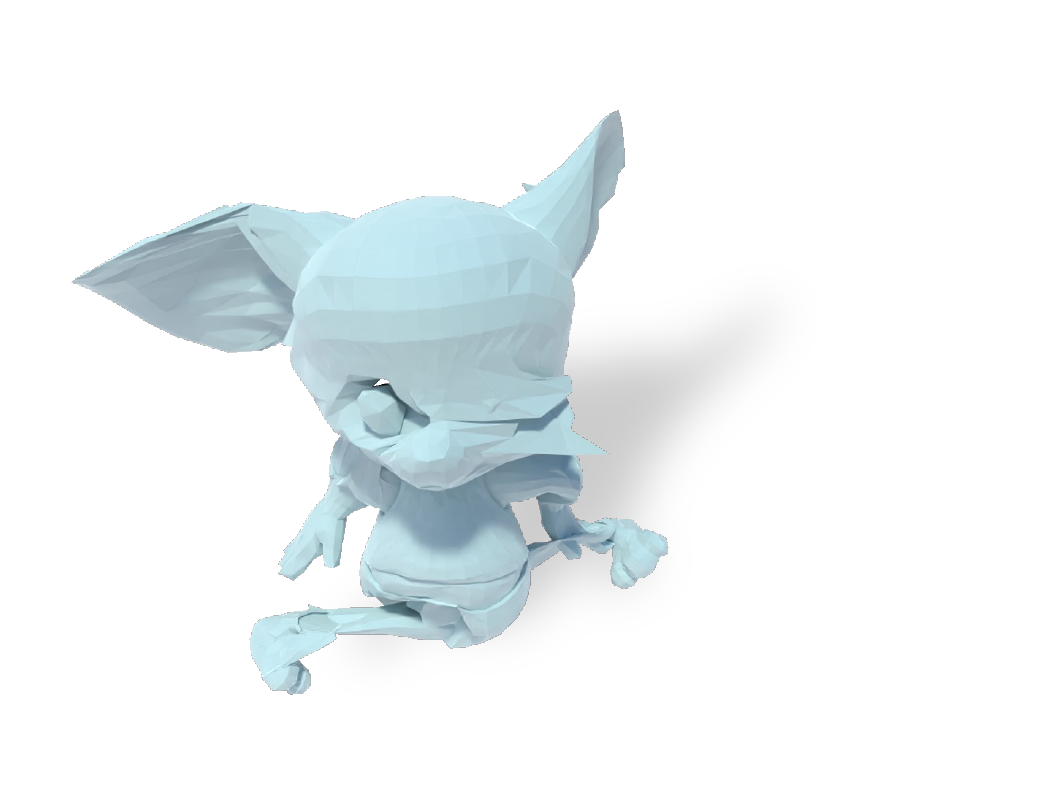}};
    \node(r2c4)[below = \vgap of r1c4]{\includegraphics[height=\imgheight, trim = {60 10 70 35}, clip]{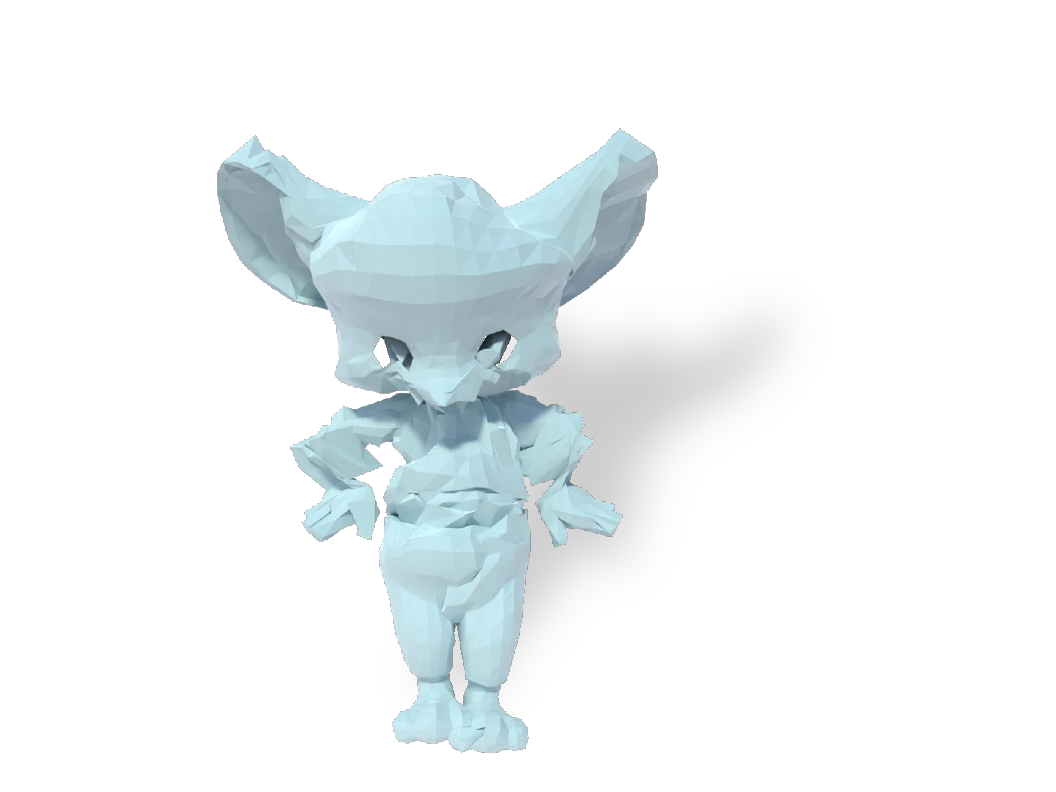}};
    \node(r2c5)[below = \vgap of r1c5]{\includegraphics[height=\imgheight, trim = {60 15 70 35}, clip]{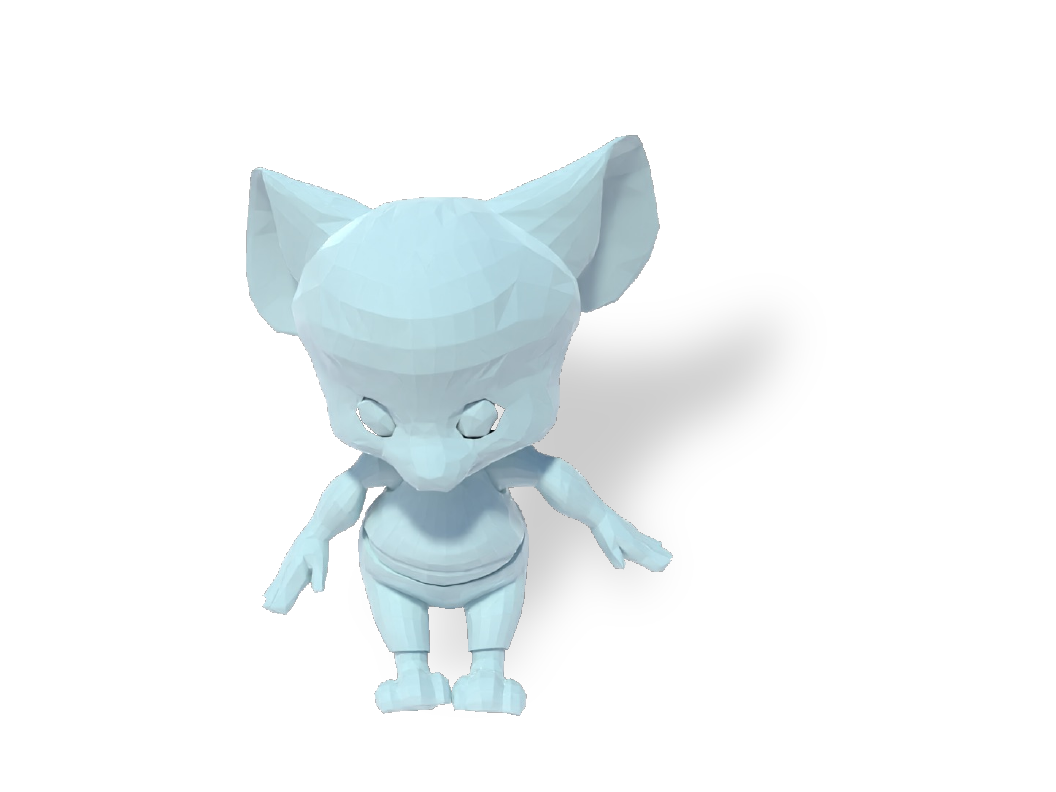}};
    \node(r2c6)[below = \vgap of r1c6]{\includegraphics[height=\imgheight, trim = {55 5 60 13}, clip]{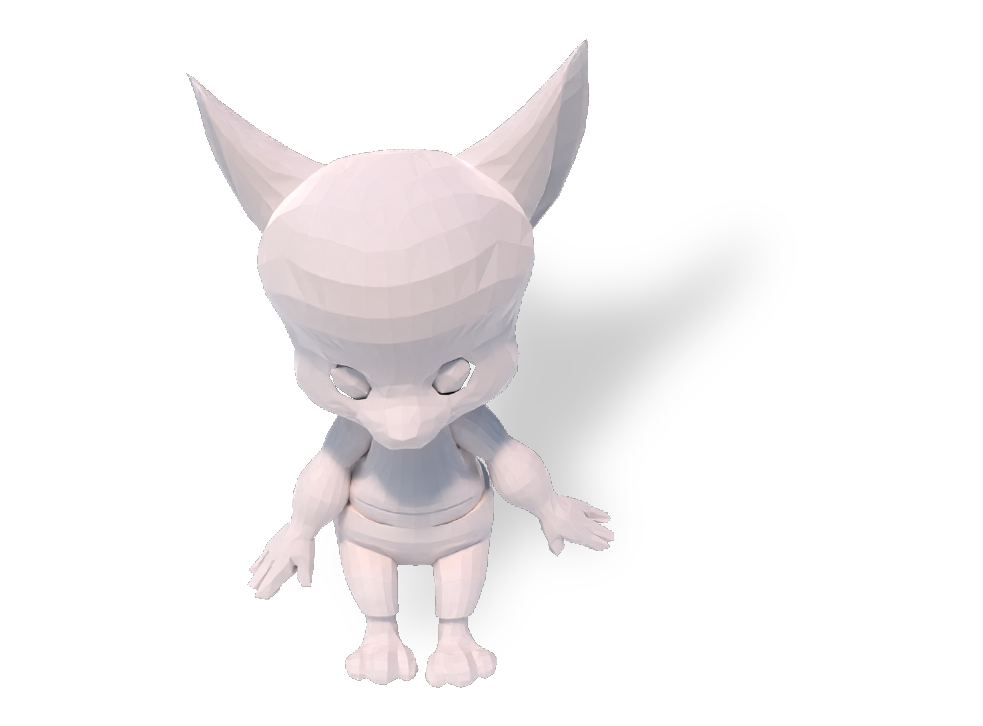}};

    \node(r3c1)[below = \vgap of r2c1]{\includegraphics[height=\imgheight, trim = {80 10 80 20}, clip]{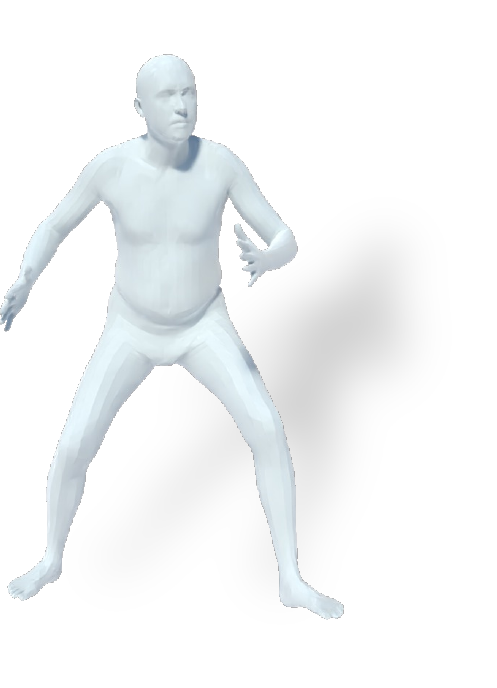}};
    \node(r3c2)[below = \vgap of r2c2]{\includegraphics[height=\imgheight, trim = {40 10 80 20}, clip]{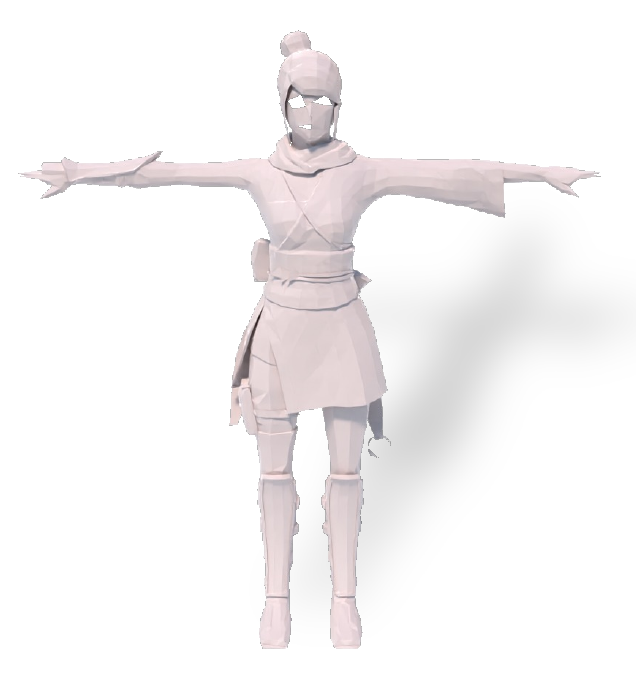}};
    \node(r3c3)[below = \vgap of r2c3]{\includegraphics[height=\imgheight, trim = {80 10 80 20}, clip]{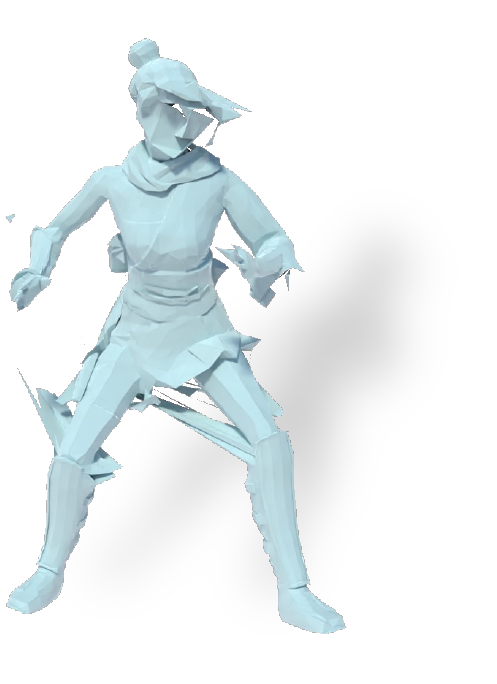}};
    \node(r3c4)[below = \vgap of r2c4]{\includegraphics[height=\imgheight, trim = {80 10 80 20}, clip]{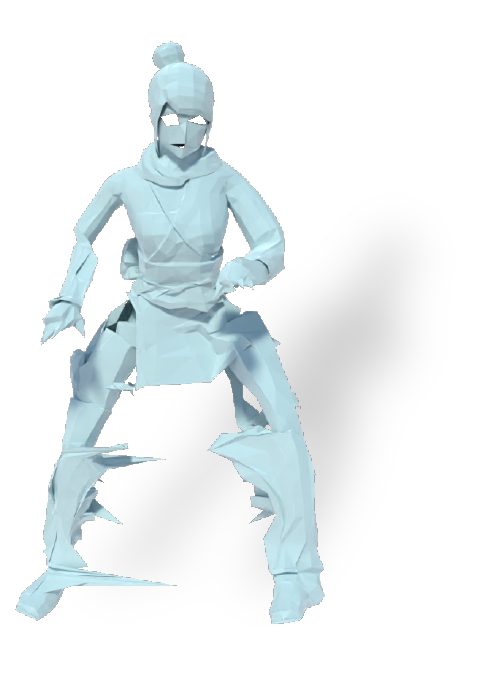}};
    \node(r3c5)[below = \vgap of r2c5]{\includegraphics[height=\imgheight, trim = {80 10 80 20}, clip]{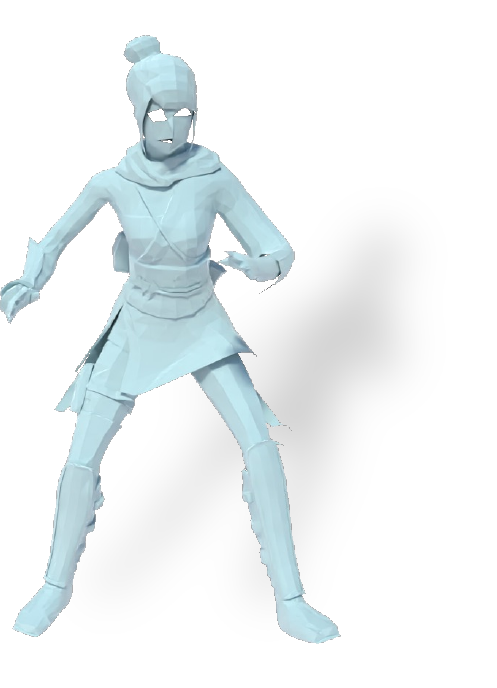}};
    \node(r3c6)[below = \vgap of r2c6]{\includegraphics[height=\imgheight, trim = {80 10 80 20}, clip]{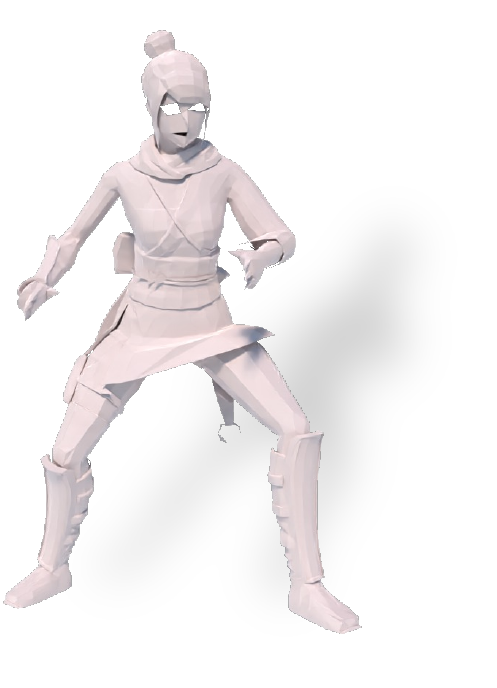}};

    \node(c1_label)[below left = 0  and 8 pt of r3c1.south, anchor = north]{Source\strut};
    \node(c2_label)[below left = 0  and 2 pt of r3c2.south, anchor = north]{Target\strut};
    \node(c3_label)[below left = 0  and 8 pt of r3c3.south, anchor = north]{NBS~\cite{li2021learning}\strut};
    \node(c4_label)[below left = 0  and 8 pt of r3c4.south, anchor = north]{SPT~\cite{liao2022skeleton}\strut};
    \node(c5_label)[below left = 0  and 8 pt of r3c5.south, anchor = north]{Ours\strut};
    \node(c6_label)[below left = 0  and 8 pt of r3c6.south, anchor = north]{GT\strut};
    


\end{tikzpicture}

%% file: 6.Conclusion.tex
\section{Conclusion}
\label{sec:conclusion}
In this paper, we present a model that deforms unrigged, stylized characters guided by a biped or quadruped avatar. Our model is trained with only easily accessible posed human or animal meshes, yet can be applied to unseen stylized characters in a zero-shot manner during inference. To this end, we draw key insights from classic mesh deformation method and develop a correspondence-aware shape understanding module, an implicit pose deformation module and a volume-based test-time training procedure. We carry out extensive experiments on both the biped and quadruped category and show that our method produces more realistic and accurate deformation compared to baselines learned with comparable or more supervision.

%% file: 8.Appendix.tex
\appendix

\section*{\Large Appendix}

In this appendix, we introduce more details about the evaluation data curation procedure, the implementation of our method and the baseline methods, more qualitative results and the limitations of our method. 

\section{Evaluation Data Curation}
\label{sec:data}
\textbf{Mixamo.} Because the preprocessed Mixamo~\cite{Mixamo} testing sequences used in~\cite{liao2022skeleton} are not publicly available, we follow the instructions in~\cite{liao2022skeleton} and download the testing data from the Mixamo website~\cite{Mixamo}. In~\cite{liao2022skeleton}, 20 stylized characters and 28 motion sequences are used for evaluation. Among the 20 characters, the ``liam'' character is not publicly available on the Mixamo website, thus we evaluate our method and the baselines on the other 19 stylized characters. 
Moreover, some evaluation motions (e.g., ``Teeter'') include more than one motion sequence on the Mixamo website with the same name. However, it is not public information as to what exact sequences were used for evaluation in the prior work~\cite{liao2022skeleton}. Thus, we download all motion sequences with the same name and randomly pick one for evaluation. %
Given a character in rest pose and the desired pose, we use the linear blend skinning algorithm to obtain the ground truth deformed mesh.
We then compare the prediction from each method with the ground truth mesh by computing the PMD and ELS scores as discussed in Sec.4.3 in the main paper.
For a fair comparison, all poses in the evaluation motion sequences are not used during training.
All methods are evaluated using these collected testing pairs.

\textbf{MGN.} We follow NBS~\cite{li2021learning} and download the MGN dataset\footnote{https://github.com/bharat-b7/MultiGarmentNetwork}, which includes 96 clothed human characters. We use the same evaluation set (i.e., the last 16 human characters) as in NBS. To obtain the ground truth deformed characters, we sample 200 poses (unseen during training) and deform each of the 16 clothed characters using the Multi-Garment Net~\cite{bhatnagar2019multi}.

\textbf{Pose code extraction from Mixamo characters.} To obtain target poses from the Mixamo motion sequences, we apply a similar fitting procedure introduced in~\cite{li2021ai}. We optimize the SMPL parameters to minimize the L2 distance between the SMPL joints and the Mixamo joints. Different from~\cite{li2021ai}, we also add a constraint to minimize the Chamfer distance between the SMPL shape vertices and the Mixamo shape vertices. 
Similarly as~\cite{rempe2021humor}, we directly optimize the pose code in the VPoser's~\cite{pavlakos2019expressive} latent space, instead of the parameters in SMPL.
We fit the SMPL shape to the "marker man" character in Mixamo to get all the testing poses.
\section{Implementation Details}
\textbf{Shape code computation.} We use an off-the-shelf method\footnote{https://github.com/marian42/mesh\_to\_sdf} that computes occupancy with ``virtual laser scans'' and does not require a watertight mesh.
We sample 10,000 points in a unit space, which takes \textbf{2.35s} on average. Then, we use the occupancy of each query point as supervision to optimize the shape code. We run 2,000 iterations with a batch size of 2,000 to get the shape code, which takes \textbf{3.41s} on average. For each character, we only compute its shape code \textbf{once} and use it to transfer poses from different motion sequences. All the time cost reported in this supplementary was measured on a laptop with I7-11700h and a RTX 3060.

\textbf{Detailed test-time training (TTT) procedure.} Following the inference procedure in~\cite{liao2022skeleton}, TTT takes a stylized character in T-pose, and a source human character in T-pose and target pose as inputs.
TTT finetunes the pose module to perform two tasks: a) the T-pose stylized character is deformed to the target pose, while being constrained by the self-supervised volume-preserving loss $L_v$. b) the source human character in T-pose is deformed to the target pose, while being supervised by the ground truth human character in the target pose ($L_{dr}$). TTT further refines the results' smoothness and resemblance to driving poses. 
$L_{dr}$ helps the pose module understand and generalize to the target pose, rather than enforcing that the human and stylized character have similar offsets.
TTT is carried out for each pair of stylized character and target pose. It is highly efficient and only requires fine-tuning the pose module for 20 iterations, which takes \textbf{18ms} without batching. We can speed it up to \textbf{12ms} for each pair with a batch size of 8.

\section{Baseline Methods Implementation}
\label{sec:implementation}
\textbf{NBS~\cite{li2021learning}.} We evaluate NBS using its publicly available code and pre-trained model\footnote{https://github.com/PeizhuoLi/neural-blend-shapes}. NBS~\cite{li2021learning} takes the SMPL pose parameters as input, thus we feed the optimized SMPL parameters discussed above to NBS.

\textbf{SPT~\cite{liao2022skeleton}.} To evaluate both SPT(full) and SPT on human-like stylized characters, we use the publicly available code\footnote{https://github.com/zycliao/skeleton-free-pose-transfer} and pre-trained models generously provided by the authors. For the quadruped category, we train and evaluate the SPT model using its public code on the dataset discussed in Sec.4.1 in the main paper. Specifically, we utilize the SMAL model~\cite{zuffi20173d} to produce motion pairs, including an animal mesh in rest pose and the desired pose. We also supervise SPT with the ground truth skinning weights from SMAL. Note that our model is trained and evaluated using the same quadruped dataset as SPT.

\section{Visualization}
\label{sec:visualization}
\begin{figure*}
  
  \vspace{-0.2in}
  \centering
  \begin{tikzpicture}
  
      \node(image) at (0,0){\includegraphics[width=0.95\textwidth,trim={0 35 0 20}, clip]{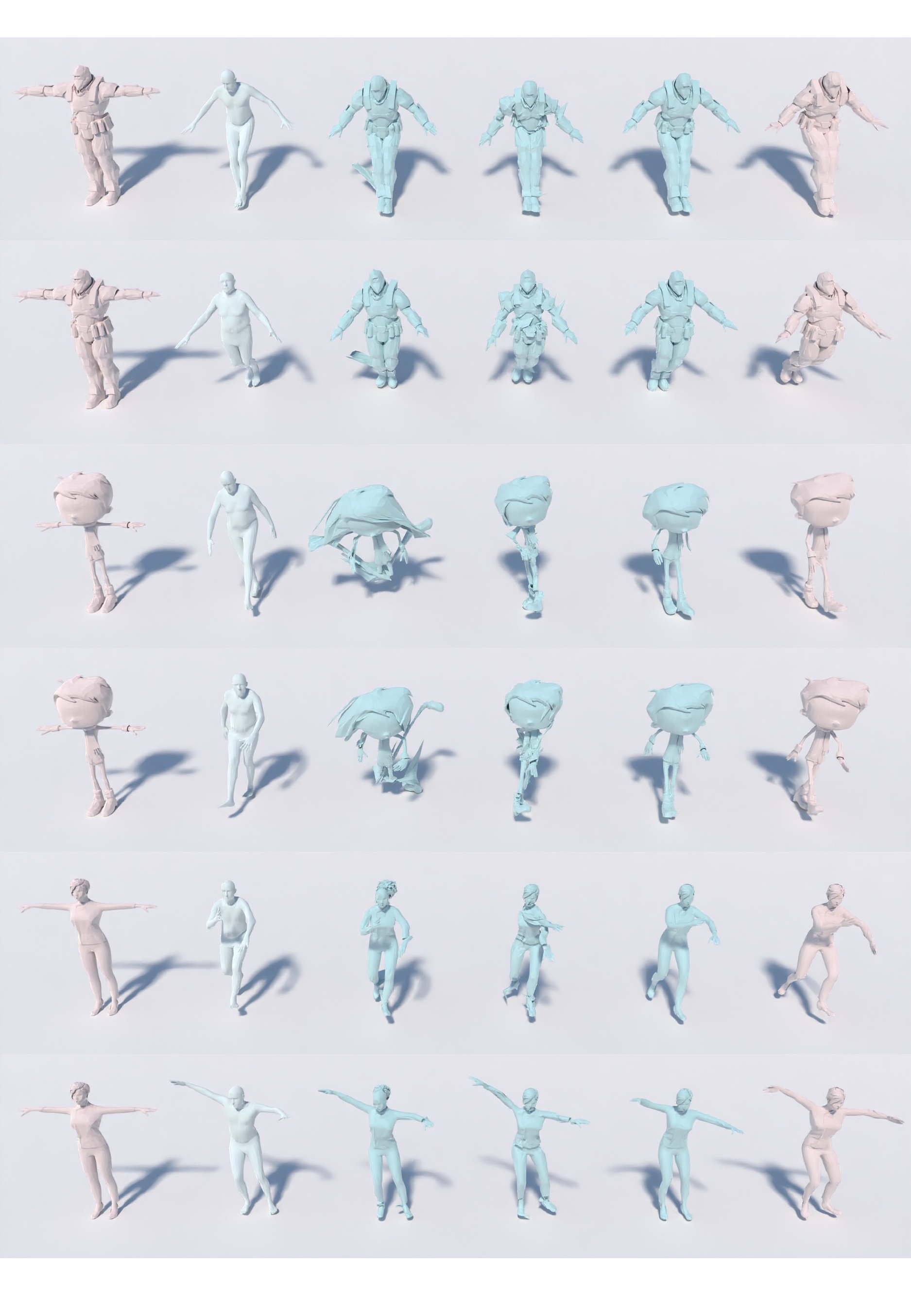}};
      \node(target_label) [below left = -3pt and 166 pt of image.south]{Target\strut};
      \node(source_label) [below left = -3pt and 87 pt of image.south]{Source\strut};
      \node(nbs_label) [below left = -3pt and 8 pt of image.south]{NBS~\cite{li2021learning}\strut};
      
      \node(spt_label)  [below right = -3pt and 25pt of image.south]{SPT~\cite{liao2022skeleton}\strut};
      \node(ours_label) [below right = -3pt and 100pt of image.south]{Ours\strut};
      \node(gt_label) [below right = -3pt and 175pt of image.south]{GT\strut};
  \end{tikzpicture}
  \vspace{-3mm}
  \caption{\textbf{Qualitative comparisons on Mixamo~\cite{Mixamo}.}}
  \label{fig::supp_compare}
\end{figure*}

\begin{figure*}
  
  \vspace{-0.2in}
  \centering
  \begin{tikzpicture}
  
      \node(image) at (0,0){\includegraphics[width=0.95\textwidth,trim={0 80 0 0}, clip]{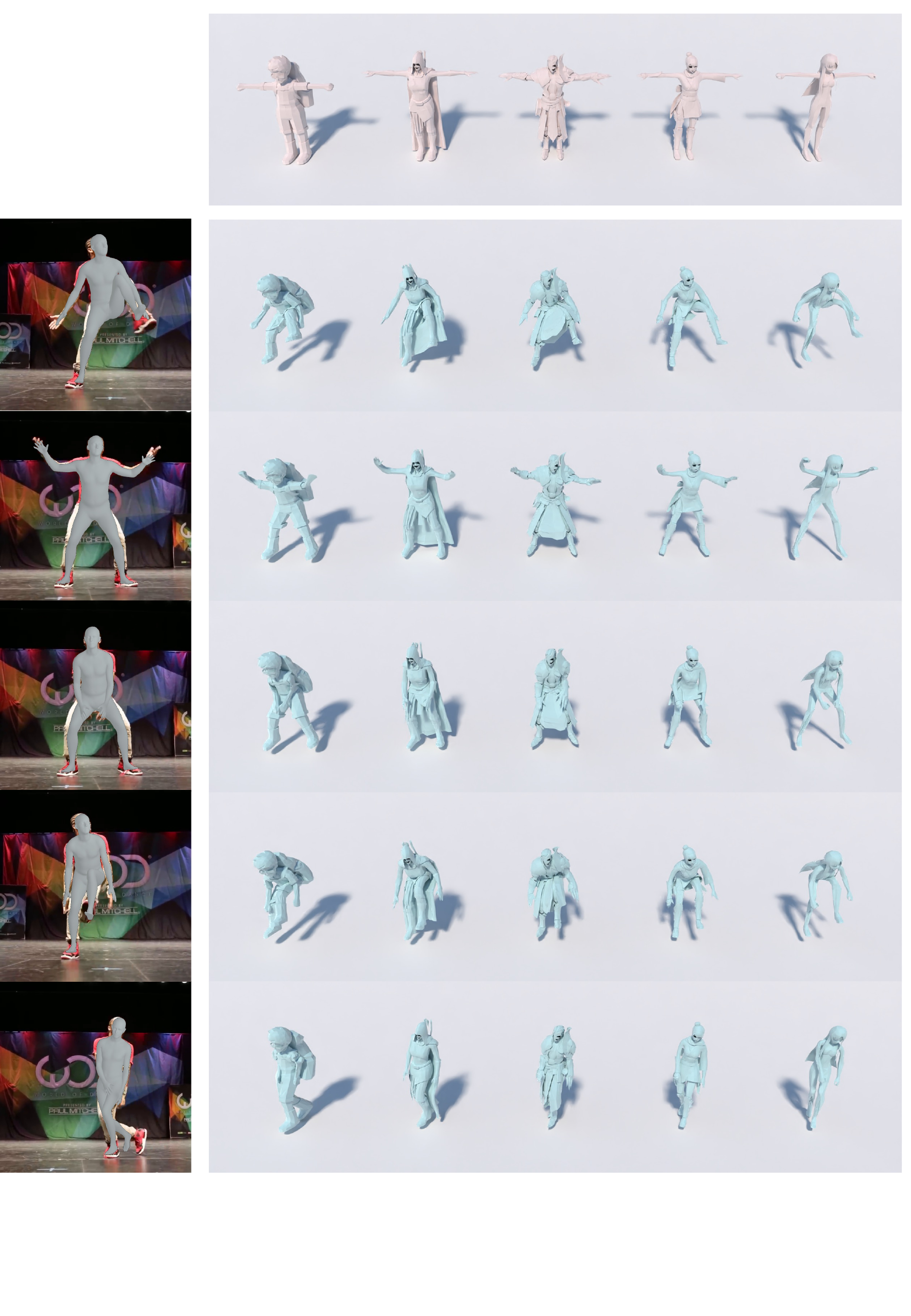}};
      \node(target_label) [below right = -10pt and 33 pt of image.north]{Target\strut};
      \node(source_label) [below left = -5pt and 166 pt of image.south]{Source\strut};
      \node(results_label) [below right = -5pt and 33 pt of image.south]{Results};
  \end{tikzpicture}
  \vspace{-3mm}
  \caption{\textbf{Transferring poses from in-the-wild videos to stylized characters.}}
  \label{fig::human_wild1}
\end{figure*}

\begin{figure*}
  \centering
  \vspace{-0.2in}
  \begin{tikzpicture}
  
      \node(image) at (0,0){\includegraphics[width=0.95\textwidth,trim={0 80 0 0}, clip]{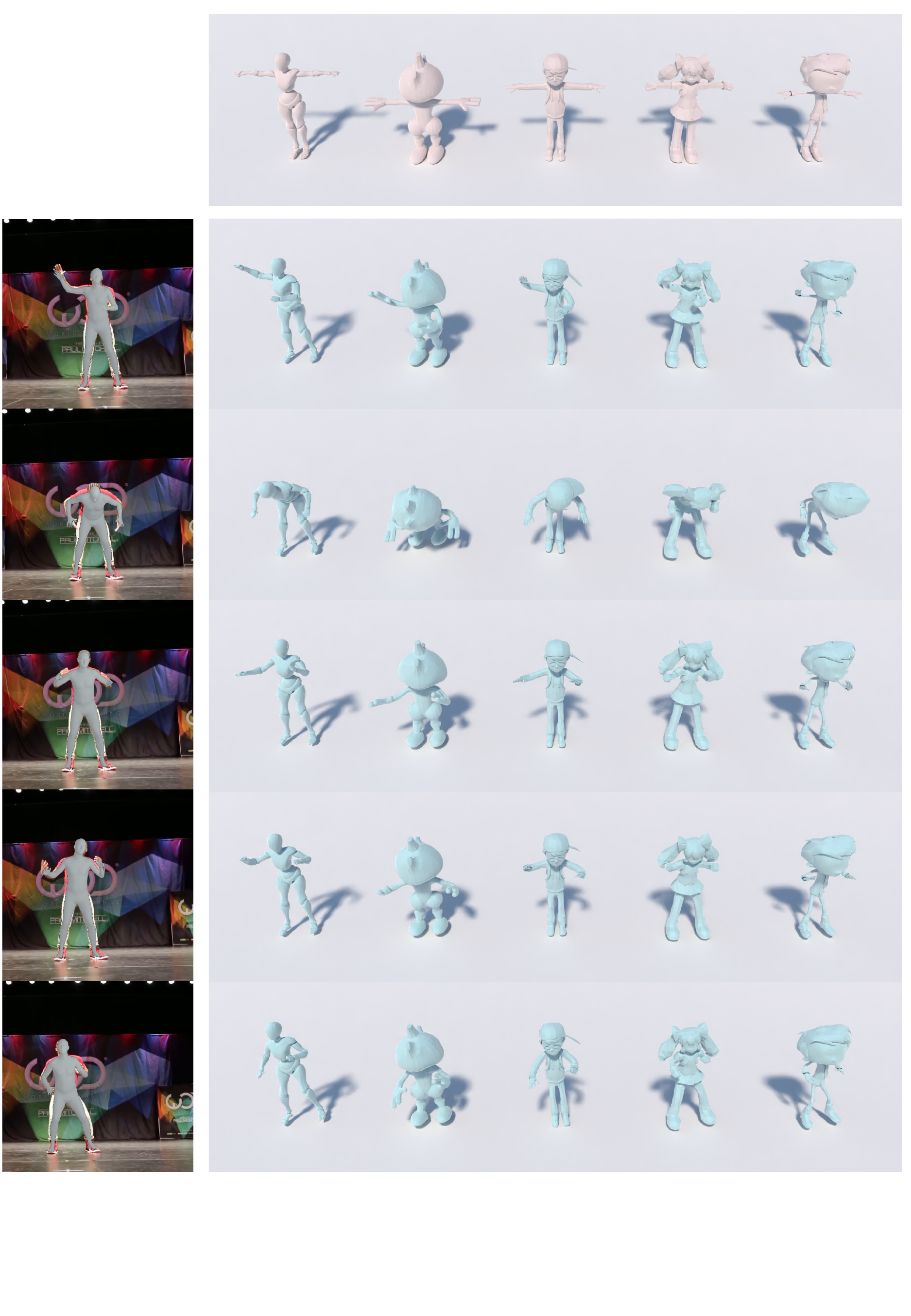}};
      \node(target_label) [below right = -10pt and 33 pt of image.north]{Target\strut};
      \node(source_label) [below left = -5pt and 166 pt of image.south]{Source\strut};
      \node(results_label) [below right = -5pt and 33 pt of image.south]{Results};
      
  \end{tikzpicture}
  
  \caption{\textbf{Transferring poses from in-the-wild videos to stylized characters.}}
  \label{fig::human_wild2}
\end{figure*}

\begin{figure*}
  \centering
  \vspace{-0.2in}
  \begin{tikzpicture}
  
      \node(image) at (0,0){\includegraphics[width=0.95\textwidth,trim={0 80 0 0}, clip]{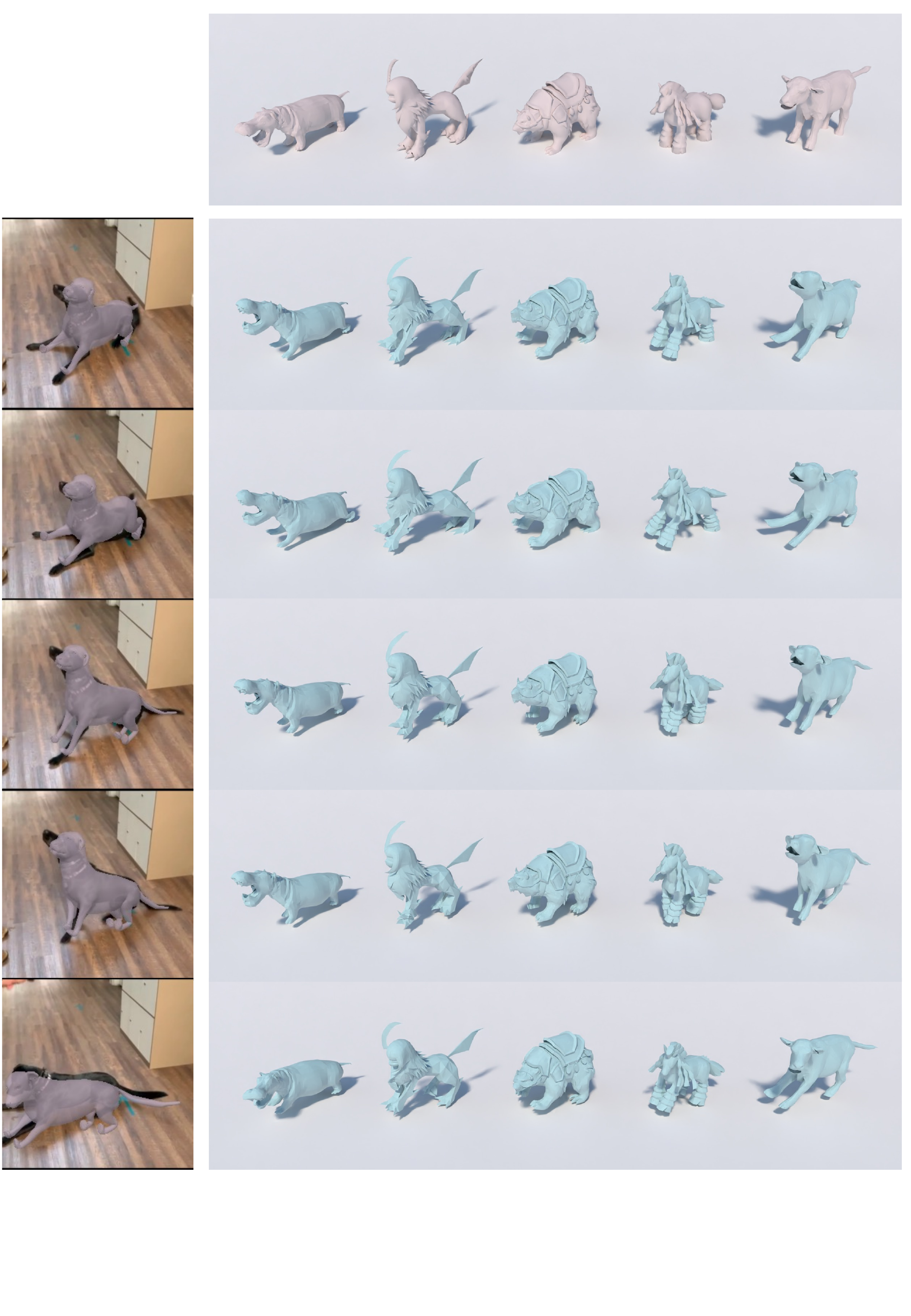}};
      \node(target_label) [below right = -10pt and 33 pt of image.north]{Target\strut};
      \node(source_label) [below left = -5pt and 166 pt of image.south]{Source\strut};
      \node(results_label) [below right = -5pt and 33 pt of image.south]{Results};
  \end{tikzpicture}
  
  \caption{\textbf{Transferring animal poses from in-the-wild videos to stylized quadrupedal characters.}}
  \label{fig::animal_wild}
\end{figure*}

We provide more visualizations, including qualitative comparisons (Fig.~\ref{fig::supp_compare}), deformation results by using source poses from in-the-wild videos for both human-like (Fig.~\ref{fig::human_wild1} and Fig.~\ref{fig::human_wild2}) and quadupeds (Fig.~\ref{fig::animal_wild}). To obtain the pose code from a video frame, we apply PyMAF~\cite{zhang2021pymaf} for human and BARC~\cite{ruegg2022barc
} for quadupeds. We provide more visualizations in the supplementary video.

\section{Limitation}
Although our approach exhibits good generalization performance for bipedal and quadrupedal characters, modeling other categories whose poses are not being studied well remains difficult. Additionally, our method is unable to solve the articulation of hands and just treats them as rigid parts. 

